\documentclass{article}

\usepackage{PRIMEarxiv}

\usepackage[utf8]{inputenc} 
\usepackage[T1]{fontenc}    
\usepackage{hyperref}       
\usepackage{url}            
\usepackage{booktabs}       
\usepackage{amsfonts}       
\usepackage{nicefrac}       
\usepackage{microtype}      
\usepackage{lipsum}
\usepackage{fancyhdr}       
\usepackage{graphicx}       
\graphicspath{{media/}}     
\usepackage{algorithm}
\usepackage{algorithmic}
\usepackage{multirow}

\title{Self-Validated Physics-Embedding Network: A General Framework for Inverse Modelling
\thanks{\textit{\underline{Update information}}: 
\textbf{Kang R. et al. updated on Octobor 17,2022 DOI:10.48550/arXiv.2210.06071}} 
}
\author{
   Ruiyuan Kang \\
  Department of Mechanical Engineering \\
  Khalifa University \\
  Abu Dhabi, UAE\\
  \texttt{ruiyuan.kang@ku.ac.ae} \\
   \And
  Dimitrios C. Kyritsis \\
  Department of Mechanical Engineering, RICH Center \\
  Khalifa University \\
   Abu Dhabi, UAE\\
  \texttt{dimitrios.kyritsis@ku.ac.ae} \\
     \And
  Panos Liatsis \\
  Department of Electrical Engineering and Computer Science \\
  Khalifa University \\
   Abu Dhabi, UAE\\
  \texttt{panos.liatsis@ku.ac.ae} \\
}

\begin{document}
\maketitle

\begin{abstract}
Physics-based inverse modeling techniques are typically restricted to particular research fields, whereas popular machine-learning-based ones are too data-dependent to guarantee the physical compatibility of the solution. In this paper, Self-Validated Physics-Embedding Network (SVPEN), a general neural network framework for inverse modeling is proposed. As its name suggests, the embedded physical forward model ensures that any solution that successfully passes its validation is physically reasonable. SVPEN operates in two modes: (a) the inverse function mode offers rapid state estimation as conventional supervised learning, and (b) the optimization mode offers a way to iteratively correct estimations that fail the validation process. Furthermore, the optimization mode provides SVPEN with reconfigurability i.e., replacing components like neural networks, physical models, and error calculations at will to solve a series of distinct inverse problems without pretraining. More than ten case studies in two highly nonlinear and entirely distinct applications—molecular absorption spectroscopy and Turbofan cycle analysis—demonstrate the generality, physical reliability, and reconfigurability of SVPEN. More importantly, SVPEN offers a solid foundation to use existing physical models within the context of AI, so as to striking a balance between data-driven and physics-driven models.
\end{abstract}


\section{Introduction}\label{1introduction}
\begin{figure}[hbt!]
\centering
\includegraphics[width=.7\textwidth]{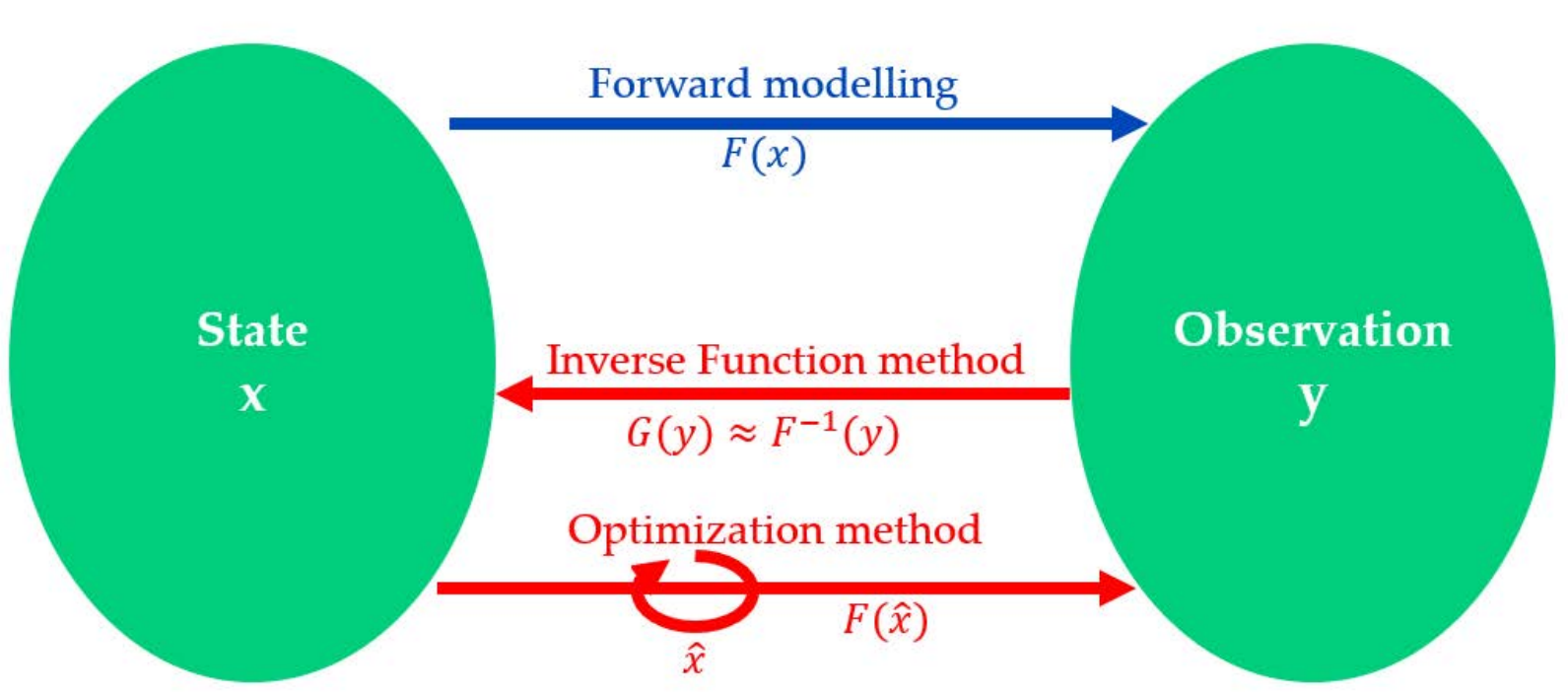}
\caption{A high-level representation of forward and inverse modelling methods}
\label{fig1}
\end{figure}

A forward problem can be defined as finding the results (observations) $y$ from a given causes (states) $x$, the solving processes which formulates the physical models $F$ to map causes (states) to are called forward modelling, i.e.,$y=F(x)$,which is demonstrated by the blue line in Fig.~\ref{fig1}. On the other hand, the counterpart problem that ascertains the states from the measured observations is named inverse problem, similarly, its solving process is named inverse modelling. Forward and inverse problems are the opposite sides of scientific and engineering problems that can be seen everywhere. For example, two forward problems can be defined as follows: (a), given the concentration and temperature of the gas cloud of a certain molecule, to calculate its absorption spectroscopy, which is often termed Molecular Absorption Spectroscopy (MAS) simulation. (b), given the cycle parameters and components parameters of an aeroengine, to calculate its performance parameters, such as thrust and thrust specific fuel consumption, which is often termed aeroengine performance simulation. Their corresponding inverse problems can be defined as: (a), from the molecular absorption spectroscopy measure, to ascertain the concentration and temperature of gas cloud of a specific molecule. (b) given the requirement of aeroengine performance, to ascertain the cycle and component parameters which satisfy the performance requirements, which is usually termed cycle analysis in aeroengine design.

In general, the methods solving the inverse problem can be categorized into two types: (a) inverse function method, as the first red line shown in Fig.~\ref{fig1}. The inverse process is modelled, so that the state corresponding to a given observation can be directly calculated, i.e., $x=G(y)\cong F^{-1}(y)$, (b) Optimization method, as the second red line shown in Fig.~\ref{fig1}. The state corresponding to the given observation is ascertained by iterating the estimated state $\hat{x}$ till a similar observation is acquired from the forward model. i.e., iterate $\hat{x}$, till $F(\hat{x})=y$. 

Substantial physics-knowledge-based solutions, inverse-function or optimization ones, have been developed for applications in specific science and engineering fields ~\cite{egbertTOPEXPOSEIDONTides1994,yehGeostatisticallyBasedInverse2002,loubetInverseModelEstimate2010,nasehitehraniL1RegularizationMethod2012,goldensteinTwocolorAbsorptionSpectroscopy2013,liewParametricCycleAnalysis2005}, however, they require substantial background knowledge and expertise. Moreover, the details of their applications constrain their generalization in other fields. 

Conversely, machine-learning-based solutions have been recently gaining popularity, due to their powerful computation capability, high flexibility, and reduced requirements for background knowledge. In the scope of inverse function method, a Machine Learning (ML) model $G(y)$ is trained to approximate the inverse function $F^{-1}(y)$ of the physical forward model ~\cite{sidkyCNNsSolveCT2021,haggstromDeepPETDeepEncoder2019,jinRecentAdvancesNeural2020,kitchinMachineLearningCatalysis2018,renThreedimensionalVectorialHolography2020,liMachinelearningReprogrammableMetasurface2019,butlerMachineLearningMolecular2018}, by using the information conveyed by a large amount of experimental or simulated data. Accordingly, for a given observation $y$, the ML model is expected to give an efficient estimation of the corresponding state. 

However, such estimations of states may be biased, since the inverse problems are usually rank-deficient ill-posed ~\cite{hansen1998rank}, which means many states exist for a specific observation. Fortunately, the forward problems are always well-posed ~\cite{yurip.WellposedIllposedIntermediate2005}. Therefore, in the scope of optimization method, ML models are usually trained to be the surrogate models $\hat{F}(x)$ of the physical forward models $F(x)$ ~\cite{conroyPhysicsEclipsingBinaries2020,huangLearningConstitutiveRelations2020,vaezinejadHybridArtificialNeural2019}, which are then used to provide an accurate estimation of the observation for the given state. Then optimization methods are utilized to iterate the state estimation $\hat{x}$ till the estimated observation $\hat{y}$ from $\hat{F}(x)$ is similar enough to the actual observation $y$. The introduction of the surrogate model brings two core benefits. First, it provides for rapid mapping from the space of states to the space of observations, which in turn support the implementation of time-consuming computations, such as Computational Fluid Dynamics (CFD) ~\cite{anDeepLearningFramework2020}. Second, the inherent differentiability properties of the network assures the appropriateness of gradient-based optimization algorithms ~\cite{xuPhysicsConstrainedLearning2022}.

The main issue with the use of the aforementioned methods is they are purely data-driven. Since there is no guarantee that the dataset reflects the true distributions of the space of states and observations, the resulting ML models cannot assure learning a sufficiently accurate mapping between these two spaces ~\cite{nandyAudacityHugeOvercoming2022}. Accordingly, ML models may not provide fairly accurate estimations for noisy data or samples outside the dataset, i.e., weak robustness ~\cite{kurakinAdversarialExamplesPhysical2017,willcoxImperativePhysicsbasedModeling2021} and generalization ~\cite{mccartneyComparisonMachineLearning2020}. The consequence is that it is not possible to guarantee the reliability of the ML estimations, thus limiting the application of purely data-driven methods for the solution of the inverse problem. 

An alleviation to this problem is injecting physical knowledge within the modelling framework. Examples of such approaches include the use of PDE-based regularization ~\cite{mengCompositeNeuralNetwork2020,tartakovskyLearningParametersConstitutive2018,karniadakisPhysicsinformedMachineLearning2021,liuSolvingInverseProblem2021} to reflect the physical conservations the model should obey, or the results from complex physical models to constrain the training direction of the model ~\cite{maGenerativeAdversarialNetworks2021,liuRethinkingBigData2016}, so that a deeper correlation between the ML and physical models can be built. Nevertheless, still no one can assure data-driven models act exactly as physical models, especially in corner cases or unseen cases. A thorough solution, which can also be regarded as injecting physical knowledge, is using the forward model of the physical process directly instead of a data-driven one, but using a network to guide the direction of state search ~\cite{maDeepFeedbackInverse2020}, as often done in meta learning ~\cite{andrychowiczLearningLearnGradient2016,finnModelagnosticMetalearningFast2017}. This kind of method is still in the scope of optimization method, If the physical model embedded is believed to be trustful, the final state estimation can thus be considered to be reliable. However, the network needs to be pretrained on the designated physical model, which means that the mismatch between the network and the physical model during testing could lead to incorrect search direction, getting stuck in local optima, and eventually cannot converge.

In summary, inverse function methods allow for efficient estimations, while optimization methods, coupled with physical forward models, provide for effective estimations. However, all these ML-based methods require pre-training, which in turn translates to requirements for preparation time and resources prior to model deployment, and more important, the concerns in regard to the generalization performance of the model on the testing dataset. 

In this work, we propose a general purpose, neural networks-based framework for the solution of inverse problems, termed Self-Validated Physics-Embedding Network (SVPEN). The principle of SVPEN is to couple both the inverse function and optimization methods, through embedding physical forward models into neural networks. By utilizing physical forward model to validate the quality of estimated state, the system ensures that final state estimations are physically reasonable, while the two problem solving modes of inverse function and optimization lend SVPEN with their advantages of efficient and effective state estimations. Moreover, SVPEN can be deployed without the use of pre-collected dataset and pre-training, while its structure can be adapted to update the underlying physical/ML models. In order to demonstrate the advantages of SVPEN, this contribution considers two diverse inverse problems as we introduced above, namely, retrieval of temperature and gas concentration from MAC, and aeroengine cycle analysis from performance requirements (code repo: \url{https://github.com/RalphKang/SVPEN_1.0}). The results demonstrate the high efficiency, effectiveness, flexibility and adaptivity of the proposed framework.

\section{SVPEN}\label{2svpen}
\begin{figure}[hbt!]
\centering
\includegraphics[width=.5\textwidth]{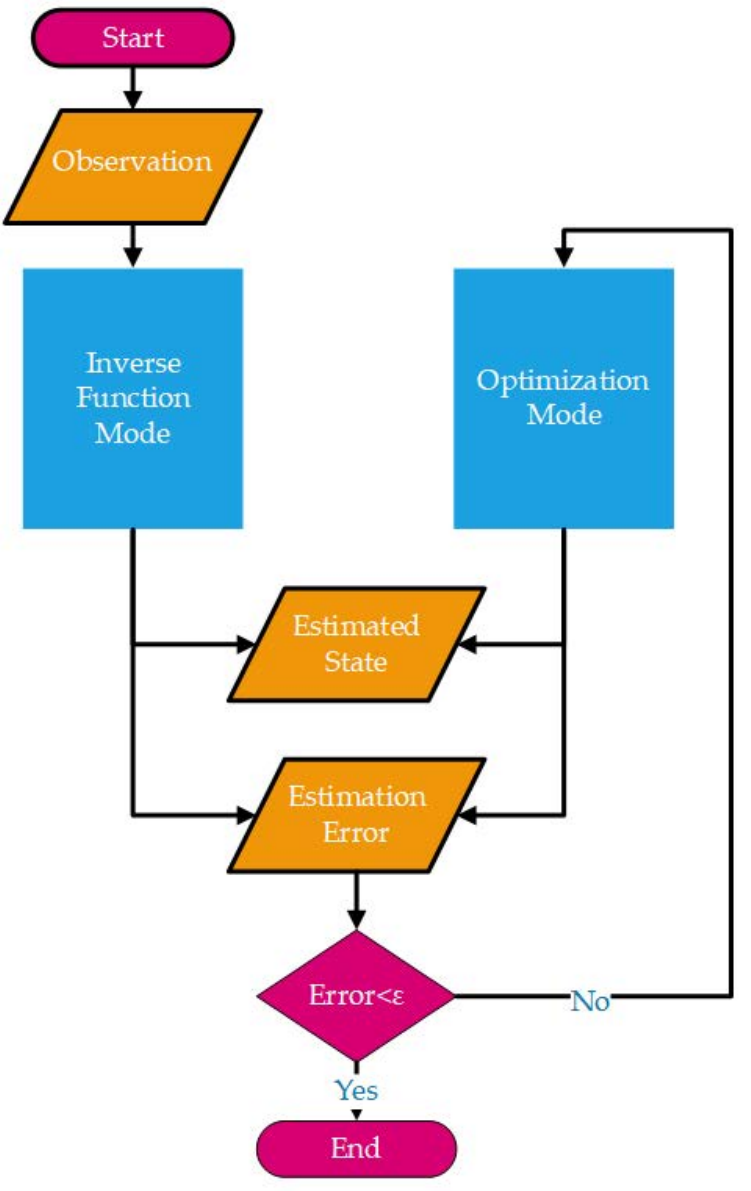}
\caption{The skeleton of SVPEN}
\label{fig2}
\end{figure}

The skeleton of Self-Validated Physics-Embedding Network (SVPEN) is demonstrated in Fig.~\ref{fig2}. In general, SVPEN is constituted of inverse function and optimization modes. Once an observation is fed to SVPEN, the inverse function mode is first used to give a quick estimation of the state; meanwhile, an estimation error which reflects the quality of state estimation is also provided. The estimated state is accepted when the estimation error is smaller than an error threshold $\varepsilon$, otherwise, the system automatically switches to the optimization mode. In the optimization mode, the estimation error is gradually reduced to be smaller than the error threshold $\varepsilon$ by performing gradient descent on the estimated state. The control variable, i.e., estimation error, is determined by the physical forward model embedded in both modes, thereby, assuring the acceptable estimated state is physically reasonable. In this section, we introduce in detail the design of SVPEN, while in section ~\ref{2.1}, we present the design of inverse function mode. In section ~\ref{2.2}, attention shifts to the introduction of the optimization mode, and finally in section ~\ref{2.3}, we explain the cooperation of these two modes within the structure of SVPEN, and the associated benefits brought about.

\subsection{Inverse function mode}\label{2.1}

\begin{figure}[hbt!]
\centering
\includegraphics[width=1.0\textwidth]{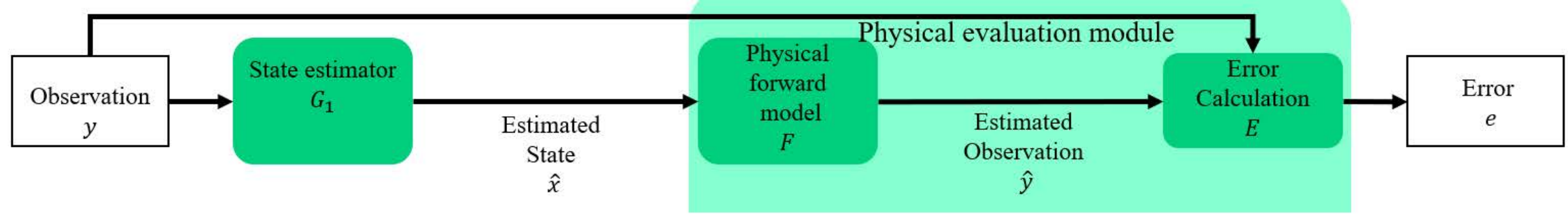}
\caption{The schematic of the inverse function mode}
\label{fig3}
\end{figure}

The structure of the inverse function mode is schematically shown in Fig.~\ref{fig3}, which is mainly constituted of three computational components, i.e., state estimator $G_1$, physical forward model $F$, and error calculation component $E$. The state estimator is a ML model, which takes the observation $y$ as the input, and transforms it to an estimated state $\hat{x}$. Then the acquired $\hat{x}$ is fed to $F$ to generate an estimated observation $\hat{y}$. The difference between estimated and given observations, and even the prior requirement of state can be used to calculate an error $e$, which assess the quality of $\hat{x}$. The whole process can be expressed into Eqs.~\ref{eq1}-~\ref{eq3}.
\begin{equation}
\label{eq1}
\hat{x}=G_1(y)
\end{equation}
\begin{equation}
\label{eq2}
\hat{y}=F(\hat{x})
\end{equation}
\begin{equation}
\label{eq3}
e=E(y, \hat{y}, \hat{x})
\end{equation}

According to the functionalities of physical forward model and error calculation component, they can be wrapped together into an integrated module termed physical evaluation module, which is used to assess the state estimation quality from a physical perspective.

\subsubsection{State estimator}\label{2.1.1}
One can recognize that the functionality of the state estimator is not different to those of ML models in the inverse function method ~\cite{sidkyCNNsSolveCT2021,haggstromDeepPETDeepEncoder2019,jinRecentAdvancesNeural2020,xuPhysicsConstrainedLearning2022,mengCompositeNeuralNetwork2020,yangGeneralFrameworkCombining2021}, i.e., it is expected to transform observations into accurate estimations of states. In order to realize such a purpose, the state estimator must be trained to learn how to map observations to states before its deployment. Similarly to the inverse function method, the training is also done in a supervised way by utilizing the observation-state pairs collected from physical experiments or simulation studies. This supervised training process is termed as pretraining herein, which in order to differentiate it from the online optimization process in optimization mode (section ~\ref{2.2}).

It is noteworthy that the state estimator must be a differentiable ML model. Indeed, this requirement is the foundation of the optimization mode, which will be illustrated in section ~\ref{2.2}. Accordingly, a neural-network-type model is used as default in SVPEN due to its natural differentiability, flexible structure, and extensive research foundation. For demonstration purposes, two architectures, i.e., VGG-13, a classical CNN architecture, and a multi-layer perceptron (MLP), are respectively used in the two application cases, presented in sections ~\ref{3.1} and ~\ref{3.2}.

\subsubsection{Physical evaluation module}\label{2.1.2}
As previously mentioned, the physical evaluation module consists of two operations. This module takes the given observation and the estimated state, and then feedbacks with a physics-determined error $e$ to assess the state estimation quality. The crucial aspect of $e$ is that it reflects the discrepancy between the given observation and the estimated observation corresponding to the estimated state. In general, only physical models of the forward process are recommended to be used to map the estimated state to its corresponding observation, since the mapping provided by such models is regarded as physically trustful. The discrepancy $e{_y}$ between the given and estimated observations can be calculated by a variety of distance functions, such as the Manhattan distance, the Euclidean distance, or a monotonic nonlinear transformation of distance functions, an example of which is used in section ~\ref{3.1.1}.

However, there are instances where only using the $e{_y}$ is insufficient, since the inverse problem is often rank-deficient and ill-posed ~\cite{hansen1998rank}, i.e., multiple estimates of state may lead to similar discrepancies that smaller than the error threshold $\varepsilon$. To tackle this issue, prior-knowledge regularization about the observations and states (shown as the dashed line in Fig.~\ref{fig3}) can be utilized as a part of $e$ to further constrain the state estimation. For example, in practice, the feasible domain of state is usually ascertained. As for the MAS respectively acquired from a flame and ordinary atmosphere, we do know the temperature retrieved from these two MAC, will fall into the range of a few thousand K and a few hundred K, respectively. Consequently, a regularization term can be set as shown in Eq.~\ref{eq4}:

\begin{equation}
\label{eq4}
e_{reg}=Max(\hat{x}-x_{max},0)+Max(x_{min}-\hat{x},0)
\end{equation}

Where, $e_{reg}$ is the regularization term, $x_{max}$ and $x_{min}$ are the upper and lower boundaries of the feasible domain of state, respectively. Therefore, the state estimation $ \hat{x}$ exceeds the feasible domain will cause extra error. Besides, the use of traditional L$_1$/L$_2$ norms to constrain the magnitude of the state, or the use of PDE-based regularizations to reflect the physical laws ~\cite{raissiPhysicsinformedNeuralNetworks2019}, etc., can also be applied, depending on the inverse problem to be solved. For instance, the total error e can be defined as Eq.~\ref{eq5}:

\begin{equation}
\label{eq5}
e=c_0e_y+\sum_{i=1}^{n}c_ie_{reg,i}
\end{equation}

where the $c$’s are the weights of the discrepancy /regularization terms. The choice of weights affects the evaluation of the estimated states. For example, for a problem where no regularizations priors are needed, all weight values except $c_0$ are set to zero. The calculation of $e$ and all individual discrepancy/regularization error terms is performed by the error calculation component $E$.

\subsection{Optimization mode}\label{2.2}
However, as we discussed above, the state estimator $G_1$ cannot be assured to always provide reasonable estimations, as it is a data-driven model. In cases where the calculated error $e$ is greater than the default error threshold $\varepsilon$, the estimated state is supposed to be far from its true state, and is thus, deemed as unacceptable. Under such conditions, SVPEN switched to the optimization mode. The purpose of the optimization mode is to utilize the backpropagation of error to optimize the estimation of state, until error $e$ satisfies the error threshold $\varepsilon$ or the number of iterations reaches the selected maximum of $t$.

\begin{figure}[hbt!]
\centering
\includegraphics[width=1.0\textwidth]{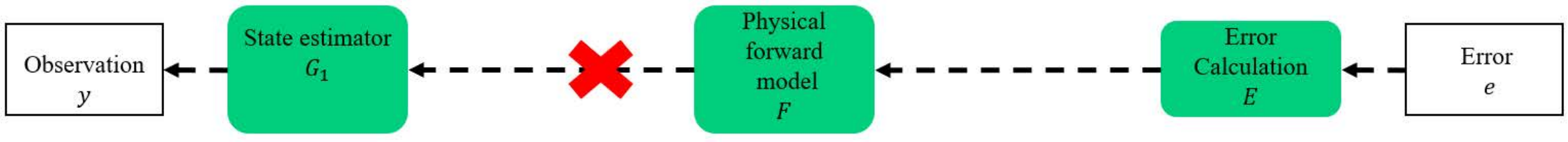}
\caption{The dilemma that  back-propagation unavailability for not differentiable physical model}
\label{fig4}
\end{figure}

However, the fact is that not all physical forward models are fully differentiable, which means that the back propagation route from the error $e$ to the state estimator $G_1$ is interrupted, as the cross symbol shown in Fig.~\ref{fig4}. Using the adjoint state method ~\cite{plessixReviewAdjointstateMethod2006} or adding perturbations to every entry of the state could support calculation of the vector of derivatives, however, these require complex transformations or frequent evaluations of the physical model in every iteration, leading to increased computational requirements. 

\begin{figure}[hbt!]
\centering
\includegraphics[width=1.0\textwidth]{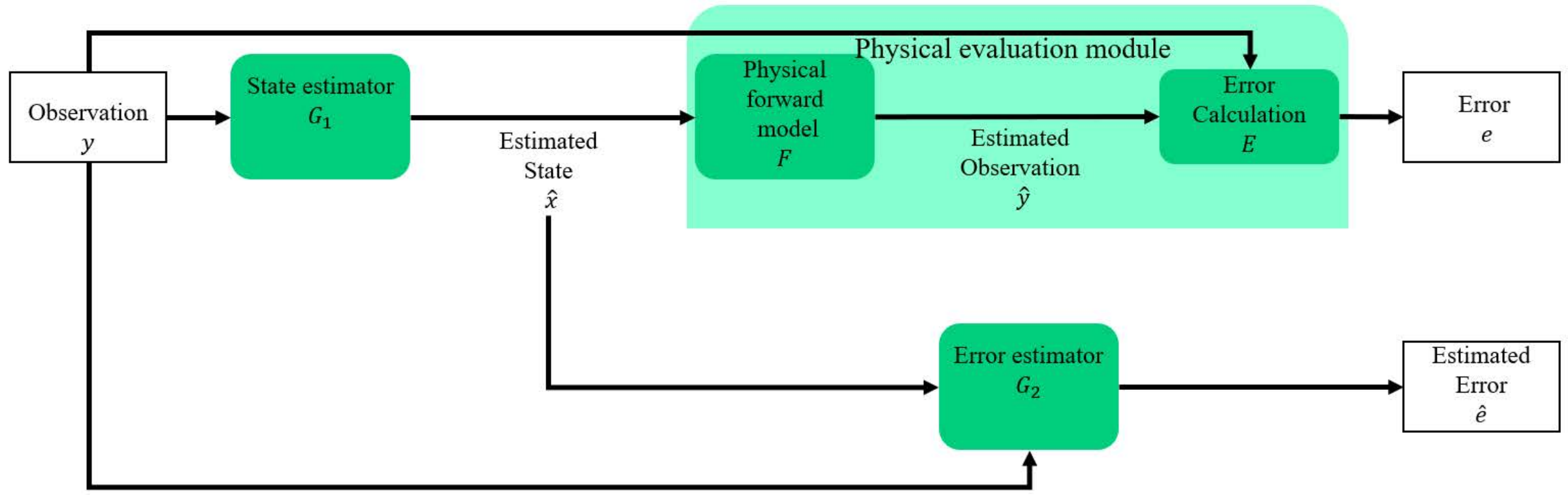}
\caption{The schematic of the optimization mode}
\label{fig5}
\end{figure}

In this research, we exploit the differentiability of the network structure of $G_1$, by bridging the error $e$ and the state estimator $G_1$ through adding another network-type learner $G_2$ to the original architecture, as shown in Fig.~\ref{fig5}. The function of $G_2$ is to estimate the error that would be provided by the physical evaluation module subject to a given observation and estimated state (Eq.~\ref{eq6}). Accordingly, $G_2$ is termed error estimator. Therefore, the error backpropagation is enabled by using the estimated error $\hat{e}$ to mimic the original error $e$.

\begin{equation}
\label{eq6}
\hat{e}=G_2(\hat{x},y)
\end{equation}

\subsubsection{Loss function and optimization strategies}\label{2.2.1}
In the optimization process, $G_2$ is first tuned in order to determine the optimization direction of state estimation. The loss function for this task can be defined as Eq.~\ref{eq7}, and the loss backpropagation is in turn used to update $G_2$. As for $G_1$, the optimization target of the state estimator is to minimize the estimated error $\hat{e}$ for a given observation $y$, and a loss function can be consequently defined as Eq.~\ref{eq8}. This loss is only used to update merely the $G_1$, although $G_2$ is used in calculation of $Loss_{G_1 }$, its learning parameters are frozen in updating.
\begin{equation}
\label{eq7}
Loss_{G_2}=|e-G_2(\hat{x},y)|
\end{equation}
\begin{equation}
\label{eq8}
Loss_{G_1}=\hat{e}=G_2(G_1(y),y)
\end{equation}

During the update of both networks, although the final target is to let $G_1$ provide accurate estimation of state, the update of $G_2$ should be emphasized since it is the driver holding the direction of updating $G_1$. Therefore, prior to $G_1$ according to $Loss_{G_1 }$, we firstly updated multiple times of $G_2$ according to$Loss_{G_2 }$, in order to ascertain a proper optimization direction, this operation is also refered as delayed update ~\cite{fujimotoAddressingFunctionApproximation2018}. Meanwhile, as slight changes of the parameters in $G_2$ may cause significant fluctuations in the updates of $G_1$, we only update $G_2$ when $Loss_{G_2 }$ is no less than a threshold of $10^{-4}$ (Eq.~\ref{eq9}) , in order to stabilize the optimization. Meanwhile, the Cosine Annealing warmup start strategy ~\cite{gotmareCloserLookDeep2018} is used as the learning rate update schedule for both $G_1$ and $G_2$, in order to avoid the two estimators to be trapped in local minima.

\begin{equation}
\label{eq9}
Loss_{G_2}\ge10^{-4}\to update  G_2
\end{equation}

\subsubsection{Architecture of error estimator}\label{2.2.2}
\begin{figure}[hbt!]
\centering
\includegraphics[width=0.8\textwidth]{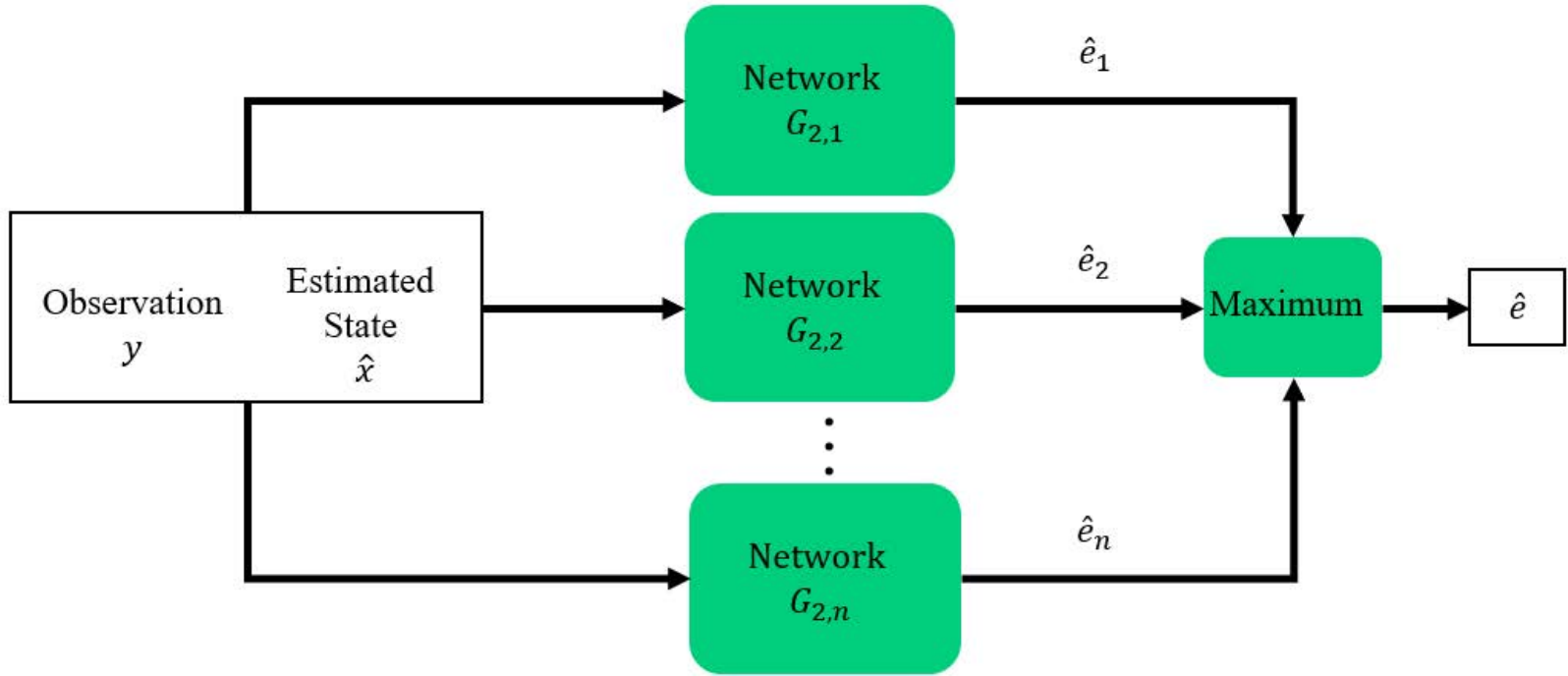}
\caption{The architecture of the error estimator}
\label{fig6}
\end{figure}

The error estimator $G_2$ tries to estimate the error from the given observation and estimated state (Eq.~\ref{eq6}). $G_2$ can be a single network, but the adversarial updating style of networks $G_1$ and $G_2$ forces $G_2$ to be prone to underestimate the error. To alleviate this condition and accelerate the convergence of SVPEN, $G_2$ can be set as an ensemble of $n$ networks instead of a single network, as shown in Fig.~\ref{fig6}. Every network in the error estimator can individually provide an estimated error, i.e., $\hat{e}_1,\hat{e}_2,\dots \hat{e}_n$ for Network $G_{2,1},G_{2,1},\dots G_{2,n},$ respectively. Due to the difference of learnable parameters or even structure of these networks, the estimated errors are not identical.  The maximum of these estimated errors is regarded to be the final estimated error $\hat{e}$ to against the underestimation of the error. Correspondingly, Eq.~\ref{eq6} can be extended to Eq.~\ref{eq10}. Albeit a “forest” of networks can be employed, we choose to use just two identical networks to construct $G_2$, so as to balance convergence speed and computation expenses.
\begin{equation}
\label{eq10}
\hat{e}=G_2(\hat{x},y)=max(G_{2,1}(\hat{x},y),G_{2,2}(\hat{x},y),\dots G_{2,n}(\hat{x},y))
\end{equation}

In principle, the network architecture in $G_2$ does not have any constraints, but since the means of calculating error utilized in our case studies (section  ~\ref{3.1} and ~\ref{3.2}) always provide non-negative values, we utilized tansig function to activate the output of each network. Albeit tansig function is nonlinear and constrains the estimated error to be between 0-1, it is worth noting that estimating the exact value of the actual error is not critical; instead, $G_2$ only needs to be sensitive to whether an estimated state should have a large/small $\hat{e}$, thus providing the appropriate step size of descent optimization. Accordingly, the bounded output range of $G_2$ is not an issue, conversely, it stabilizes the optimization of $G_2$. To remedy the relatively narrow range of tansig function, a scaling factor $s>1$ can be multiplied with the output of tansig function to amplify the range of values. Although ReLU seems to be a more elegant option than tansig function due to its linearity and simpleness, from our practice, it eliminates negative outputs of the network, which slows down the optimization process.

\subsubsection{Utilization of replay buffer}\label{2.2.3}
Although merely using the current estimate of state $\hat{x}_c$ and its corresponding error $\hat{e}_c$ can enable the optimization of $G_2$, such a type of stochastic learning is unstable and can easily converge into local optima. In order to balance the exploration and exploitation ~\cite{ishiiControlExploitationExploration2002}, rather than only using current estimates, we also utilized samples from two replay buffers, i.e., the in-situ exploration buffer $B_{IE}$ and the random walk buffer $B_{RW}$, to accelerate and stabilize the optimization process. The in-situ exploration buffer stores noise added estimated states $x_{noise}$ and their corresponding errors $e_{noise}$, the operation can be described by Eqs.~\ref{eq11}-~\ref{eq13}.
\begin{equation}
\label{eq11}
x_{noise}=\hat{x}_c +\varsigma_{Gaussian}
\end{equation}
\begin{equation}
\label{eq12}
e_{noise}=E(y,F(x_{noise}), x_{noise})
\end{equation}
\begin{equation}
\label{eq13}
B_{IE} \gets (x_{noise},e_{noise})
\end{equation}
Where, $\hat{x}_c$, $\varsigma_{Gaussian}$ are respectively current estimated state and gaussian noise.

The random work buffer stores random assigned states $x_r$ and its corresponding errors $e_r$, as described by Eqs.~\ref{eq14}-~\ref{eq16}.One can see from Eq.~\ref{eq14}, we also utilized feasible domain, if available, to constrain the random sampling of $x_r$.
\begin{equation}
\label{eq14}
x_{r}=random(x_{min},x_{max})
\end{equation}
\begin{equation}
\label{eq15}
e_{r}=E(y,F(x_{r}), x_{r})
\end{equation}
\begin{equation}
\label{eq16}
B_{RW}\gets (x_{r},e_{r})
\end{equation}

To some extent, the in-situ exploration buffer and random work buffer can be analogous to the fine and coarse focus functions of a microscope, respectively. These help $G_2$ to probe the properties of the regional and global location of the current estimation so as to determine the appropriate direction for gradient descent.

In each epoch of the optimization mode, a batch of estimated states $\hat{x}_B$ and their corresponding errors $e_B$ are utilized (Eq.~\ref{eq17}), and the loss function Eq.~\ref{eq7} is extended to Eq. ~\ref{eq18}.
\begin{equation}
\label{eq17}
y, \hat{x}_B, e_B \gets \left\{ B_{RW},B_{IE}, (y,\hat{x}_c,e_c) \right\}
\end{equation}
\begin{equation}
\label{eq18}
Loss_{G_2}=\frac{\sum_{i=1}^{d}(e_{B,i}-G_2(\hat{x}_{B,i},y))^2}{d}
\end{equation}
Where, $d$ is the amount of samples in one batch.As elaborated by Eqs.~\ref{eq17}, the data batch is constituted by the samples from both buffers and current estimations. In practice, the batch includes eight samples respectively supplied by each of the two buffers, and one sample from the current estimation, thus $d$ = 17. Indeed, the composition and the size of the batch can be adjusted. For convenience, the random walk buffer first runs sixteen times to collect the necessary amount of data to activate the first iteration. In each iteration, in addition to using the physical model to calculate the error for the current estimated state, two additional runs of the physical model are utilized to collect the data for the in-situ exploration buffer and the random walk buffer, respectively. From the second epoch of the optimization mode onwards, the sample amount extracted from the in-situ exploration buffer gradually increases until it reaches eight samples, while the sample amount from the random walk buffer correspondingly decreases to eight samples.

\subsubsection{Pretraining of error estimator}\label{2.2.4}
\begin{figure}[hbt!]
\centering
\includegraphics[width=.7\textwidth]{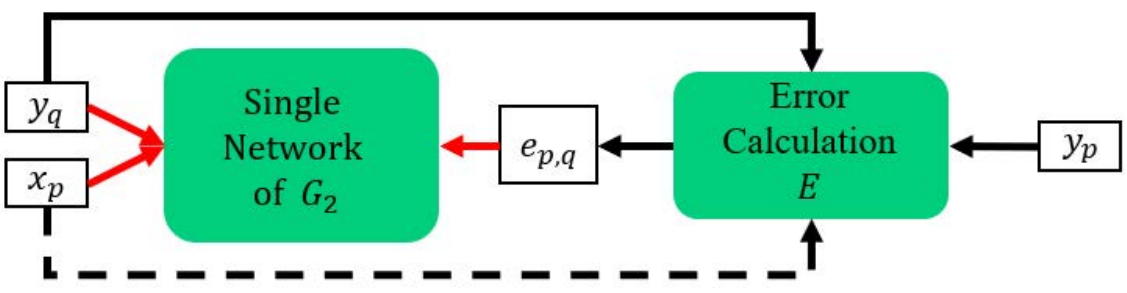}
\caption{Pretraining mechanism of $G_2$}
\label{fig7}
\end{figure}

In addition to online update of $G_2$, pretraining is also available for $G_2$, which can help $G_2$ build the awareness of the relationship between the given observation and the estimated state in advance, thus accelerate the optimization process of ascertaining state. The mechanism of pretraining $G_2$ is shown in Fig.~\ref{fig7}. In such a pretraining, a combination of state-observation variables from the dataset for training $G_1$ are utilized as inputs for the networks in $G_2$. The fed state $x_p$ and observation $y_q$ are not limited to be paired, and instead, it is better they are mainly unpaired data, i.e., $p \not\equiv q$, so that to let $G_2$ sense the error. The corresponding errors calculated from the error calculation component are utilized as labels to tune  $G_2$. The loss function for such a pretraining can be described by Eq.~\ref{eq19}.This kind of pretraining can be regarded as an Energy-based model ~\cite{lecunEnergyBasedModelsDocument2007} or variant of contrastive learning ~\cite{khoslaSupervisedContrastiveLearning}.
\begin{equation}
\label{eq19}
Loss_{G_2}=\frac{\sum_{i=1}^{d}(E(x_{p,i},y_{q,i},y_{p,i})-G_2(x_{p,i},y_{q,i}))^2}{d}
\end{equation}

 It is worth emphasizing that this pretraining, i.e., utilizing prepared dataset to do supervised learning, is optional, since (1) online optimization can solve the problem individually. (2) when utilizing the reconfigurability of SVPEN, pretrained $G_2$ and physical evaluation module are often mismatch, this pretrain is meaningless. In practice, the two examples demonstrated in sections ~\ref{3.1} and ~\ref{3.2} do not employ such a procedure.

\subsection{Summary of SVPEN}\label{2.3}
The pseudo code of SVPEN is presented in Algorithm ~\ref{algorithm1}. The workflow of SVPEN can be summarized as follows: For a given observaton $y$, the inverse function mode is first activated, the state estimator $G_1$  offers an efficient first attempt in state estimation, the estimated state $\hat{x}$ is accepted if it satisfied the validation from physical evaluation module, i.e., the error $e$ smaller than the threshold $\varepsilon$. Otherwise, the optimization mode starts, and the error estimator $G_2$ guide the optimization of state estimation. The incorporation of the physical forward model $F$, providing the only actual ground truth in the system, ensures that all estimated states satisfying the error threshold $\varepsilon$, are both accurate and physically reasonable. As shown in line 24 of Algorithm ~\ref{algorithm1}, the inverse function and optimization modes are only functionally rather than architecturally separated. Instead, the optimization mode reuses all components of the inverse function mode, which makes SVPEN compact.

\begin{algorithm}
\caption{SVPEN}
\label{algorithm1}
\begin{algorithmic}[1] 
\REQUIRE given observation $y$, error threshold $\varepsilon$,maximum iterations $t$ 
\STATE \textbf{Inverse Function mode starts}
\STATE $\hat{x}_c=G_1(y)$ \qquad \% Eq.~\ref{eq1}
\STATE $\hat{y}_c=F(\hat{x}_c)$ \qquad \% Eq.~\ref{eq2}
\STATE $e_c=E(y,\hat{y}_c,\hat{x}_c)$ \qquad \% Eq.~\ref{eq3}
\IF {$e_c<\varepsilon$}
\STATE $\hat{x}_{converge}=\hat{x}_c$
\ELSE 
\STATE \textbf{Optimization mode starts}
\STATE  Initialize $B_{RW},B_{IE}$, set $epoch=0$
\FOR{$epoch<t$}
\STATE sample $y,\hat{x}_B,e_B \gets \left\{ B_{RW},B_{IE}, (y,\hat{x}_c,e_c)\right\}$ \qquad \% Eq.~\ref{eq17}
\STATE set $delay=0$
\FOR{$delay\leq 5$}
\STATE $Loss_{G_2}=\frac{\sum_{i=1}^{d}(e_{B,i}-G_2(\hat{x}_{B,i},y))^2}{d} $ \qquad \% Eq.~\ref{eq18}
\IF{$Loss_{G_2}>10^{-4}$} 
\STATE update $G_2$  \qquad \% Eq.~\ref{eq9}
\ENDIF
\STATE $delay=delay+1$
\ENDFOR
\STATE $Loss_{G_1}=\hat{e}=G_2(G_1(y))$ \qquad \% Eq.~\ref{eq8}
\STATE update $G_1$
\STATE $B_{IE} \gets (x_{noise},e_{noise})$: $x_{noise}=\hat{x}_c+\varsigma_{Gaussian}$; $e_{noise}=E(y,F(x_{noise}),x_{noise})$ \qquad \% Eqs.~\ref{eq11}-~\ref{eq13}
\STATE $B_{RW} \gets (X_r,e_r)$: $x_r=random(x_{min}, x_{max})$; $e_r=E(y,F(x_r),x_r)$ \qquad \% Eqs.~\ref{eq14}-~\ref{eq16}
\STATE Repeat step 2-4
\STATE $epoch=epoch+1$
\IF{$e_c<\varepsilon$}
\STATE $\hat{x}_{converge}=\hat{x}_c$, terminate Optimization mode
\ENDIF
\ENDFOR
\ENDIF
\end{algorithmic}
\end{algorithm}

\begin{figure}[hbt!]
\centering
\includegraphics[width=1.0\textwidth]{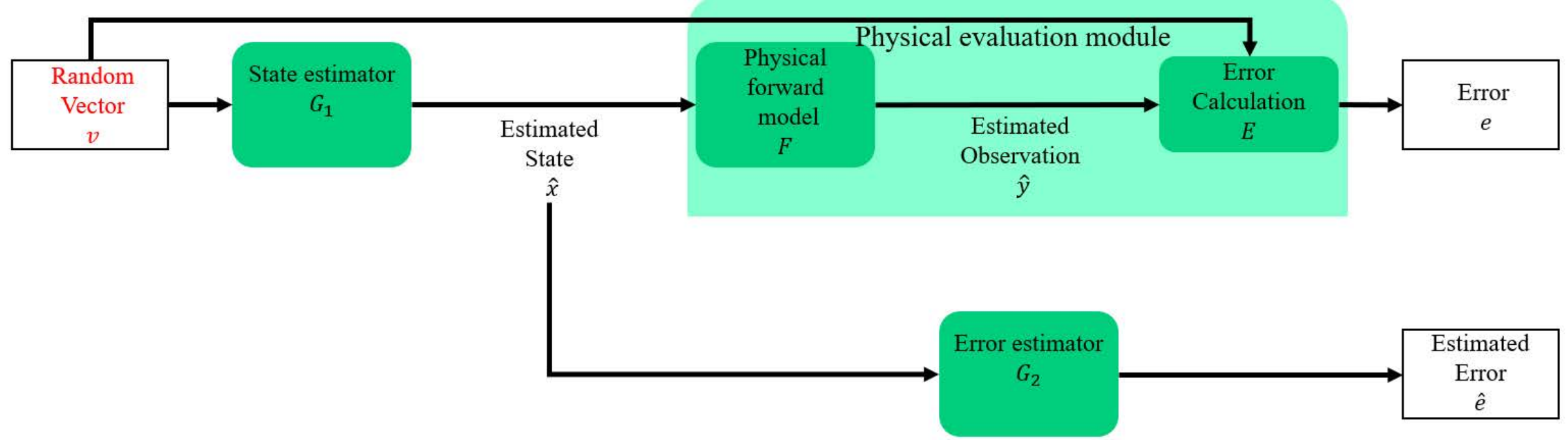}
\caption{The schematic of the expanded optimization mode}
\label{fig8}
\end{figure}

It is notable that, during optimization mode, $G_1$ and$G_2$ are optimized simultaneously by only using the feedback from the physical evaluation module. Therefore, in this mode, there is no need for neither pretraining nor dataset preparation. This mechanism brings unprecedented flexibility and adaptability to the framework, and the corresponding benefits can be listed as follows:
\begin{itemize}
	\item Without considering the accuracy of the inverse function mode, $G_1$ and $G_2$ inside the system framework can be completely free of pretraining, and merely optimization mode can find a proper estimation of state.
	\item The coupling between $G_1$,$G_2$, and the physical evaluation module is effectively relaxed, so that they can be updated/replaced at will. In turn, this leads to a reconfigurable model, which can be used in the solution of diverse inverse problems by merely replacing the associated physical evaluation module. 
	\item With a proper design of error function, The input of $G_1$ can be any random vector rather than the exact observation, and optimization mode can still approximate the state which satisfies the error threshold (Fig.~\ref{fig8}). The reason behind is, with a well-defined error function, the fully differential structure of  $G_1$ and $G_2$ will find the suitable state according to the guide of gradient backpropagation, in order to minimize error function. Such an expanded optimization mode has the benefit that relaxing the necessity of measuring exact observation, thus, allows the use of SVPEN to applications where the exact observation is unknown but several properties of the ideal observation are known, an example is given in section ~\ref{3.2}.
\end{itemize}

It is worth emphasizing that the proposed optimization mode has a similar architecture to Generative Adversarial Networks (GAN) ~\cite{creswellGenerativeAdversarialNetworks2018}, and the Actor-Critic type Reinforcement Learning (RL) algorithm ~\cite{haarnojaSoftActorCriticAlgorithms2018}. We do appreciate we learned a lot from both architectures, as one can see from the description of SVPEN above, but still, SVPEN has intrinsic difference from them. When it comes to GAN, $G_1$ and $G_2$ can be thought of as being analogous to the generator and discriminator in GAN, respectively. Albeit GAN has already been used in the context of inverse problems ~\cite{yangGeneralFrameworkCombining2021,xieSeismicInverseModeling2021,sanchez-lengelingInverseMolecularDesign2018}, the utilization style of SVPEN is thoroughly different: (a) The generator in GAN performs the mapping from the observation or noise vector to the estimated observation directly, without the use of the physical forward model which is the core of FVPEN. (b) The discriminator contrasts the difference between the fake generation (reconstructed observation) to ground truth (actual observation) with all means, whereas $G_2$ only tries to mimic the error which has exact physical meaning. We admit that SVPEN is even more similar to Actor-Critic algorithm, where $G_1$, $G_2$ and even physical evaluation module can be thought of being analogous to the actor, critic and environment in RL, respectively. Still, clear differences exist between Actor-Critic algorithm and SVPEN: (a) no target networks of actor and critic are needed to stabilize the optimization process. (b) the static optimization process in SVPEN allows the pretraining of $G_1$, so that the inverse function mode, which is vital in the framework, can be realized. (c) With proper error design, the observation can be any random vector rather than an exact observation, however, which is unacceptable for an actor in reinforcement learning.(d) Last but not least, the emphasis of optimization mode in SVPEN is to find a solution for a given observation, i.e., optimization, whereas, the emphasis of Actor-Critic is training a intelligent strategist /controller, i.e., acquiring a well-trained network.

\section{Applications}\label{3applications}
In order to demonstrate the characteristics and capability of SVPEN, we applied SVPEN to two carefully selected applications: (a) Estimating temperature and concentration of gas from Molecular Absorption Spectroscopy (MAS), and (b) Turbofan cycle analysis (TCA) according to specific performance requirements. 
The reasons of choosing these two applications are as follows: 
\begin{itemize}
    \item 	These two applications are not only totally distinctive in terms of physical principles but also somewhat opposite in terms of the posing of the associated mathematical problem, which, in turn, supports the general-purpose application of SVPEN. Specifically, gas quantification via MAS is a quantum-mechanics-based microscale problem, while TCA is an aerodynamic-thermodynamic-coupling macroscale problem. Moreover, gas quantification from MAS is a rank-redundant problem since the observation spectrum has a much higher rank than that of the state of temperature and concentration, whereas TCA is a rank-deficient problem: the observation of performance has a rank of two while the state, i.e., cycle and component parameters, has a rank of eleven. 
	\item  Both problems are complex and highly nonlinear, which will be demonstrated by their physical models in their respective sections, so that applying SVPEN to such problems can shed light into its capabilities. 
	\item Gas quantification via MAS is a perfect platform for demonstrating the reconfigurability power of SVPEN. MAS has multiple selectable wavebands to utilize, and its brother technology, i.e., emission spectroscopy, exists. Therefore, by applying SVPEN to find the states corresponding to spectra of different wavebands or even from emission, but without repeating the pretraining of $G_1$, the reconfigurability of SVPEN can be exhibited.
	\item On the other hand, the acceptable solution of TCA is somewhat infinite, the optimal solution is changeable for different turbofan performance requirement, so that it does not have the ground truth to be used in pretraining $G_1$. This is an interesting problem posing which can be used to demonstrate how to merely use the (expanded) optimization mode to acquire physically reasonable state estimation. 
	\item Last but not least, these two inverse problems are meaningful in terms of engineering applications. Gas quantification from MAS is a popular tool utilized in emission monitoring ~\cite{liuLaserAbsorptionSpectroscopy2019}, combustion diagnostics~\cite{kangEmissionQuantificationPassive2022} , etc. Cycle analysis is a common technology used not only in the design of turbofan, but also in all kinds of design of gas-turbine-based engines, e.g., those installed on ships ~\cite{altosolePerformanceDecayAnalysis2014} and powerplants ~\cite{pilavachiPowerGenerationGas2000}.
\end{itemize}
\subsection{Application I: Gas quantification from MAS}\label{3.1}
Both the simulation and inference of MAS are highly nonlinear problem. At its simplest level, the forward process, formulation of MAS, can be regarded as a transformation from a pair of given temperature $T$ and concentration $X$ , i.e., the state, to the absorptivity $\alpha$ at a bunch of wavelengths, i.e., the observation, which can be simply described as Eq.~\ref{eq20}. However, its internal transformation is sophisticated. The absorptivity at a specific wavelength can be calculated by Beer-Lambert law (Eq.~\ref{eq21}) ~\cite{hansonSpectroscopyOpticalDiagnostics2016}.
\begin{equation}
\label{eq20}
F(T,X)=\alpha
\end{equation}
\begin{equation}
\label{eq21}
\alpha _v=1-exp(-k_vl)
\end{equation}

Where, $k_v$ and $l$ are the absorption coefficient and the light path length, respectively, and the subscript $v$ represents the wavelength. For a given molecule, the absorption coefficient $k_v$  can be calculated by Eq.~\ref{eq22}. ~\cite{hansonSpectroscopyOpticalDiagnostics2016}: 
\begin{equation}
\label{eq22}
k_v=s_v(T)\phi(T,P)\frac{PX}{k_BT}
\end{equation}

where $s$ is the line intensity per molecule ($cm^{-1}/(molecule*cm^{-2})$), which is a function of temperature $T$. $\phi_v$ is the line-shape function ($cm^{-1}$), which broadens the peak intensity into a wide distribution of intensity. Such a phenomenon is called line broaden, and is controlled by temperature $T$ and pressure $P$ ~\cite{whitingEmpiricalApproximationVoigt1968}. $k_B$ is the Boltzmann constant. when looking into more details, $s$ can be regarded as a nonlinear transformation of temperature (Eq.~\ref{eq23})~\cite{hansonSpectroscopyOpticalDiagnostics2016}.
\begin{equation}
\label{eq23}
s_v(T)=s_v(T_0)\frac{Q(T_0)T_0exp(\frac{hcE^{''}}{k_BT})[1-exp(\frac{hcv_0}{k_BT})]}{Q(T)Texp(\frac{hcE^{''}}{k_BT_0})[1-exp(\frac{hcv_0}{k_BT_0})]}
\end{equation}
where $Q, h, c, E^{''},T_0$ are respectively the partition function, Plank constant, local light speed, lower state energy and reference temperature. Meanwhile, the positions of these spectral intensity lines are changed according to the temperature and pressure of gas, which is known as the line shift ~\cite{hansonSpectroscopyOpticalDiagnostics2016}.

As for the inverse problem, the inputs and outputs of the mapping function are exchanged, i.e., from the observed absorption spectrum $\alpha$ at a waveband, we need to retrieve the temperature and concentration of the molecule of interest. According to the complexity of forward problem, it is tough to derive a direct inverse function ~\cite{liuInverseRadiationProblem2019}, and the same holds for finding the optimization direction ~\cite{niuAlgorithmsRemoteQuantification2012}. A clever and practical method to find the temperature is two-color pyrometry ~\cite{goldensteinTwocolorAbsorptionSpectroscopy2013,hansonTemperatureMeasurementTechnique1978}. In this method, the temperature is measured by the ratio between the intensities of two peaks, since such a ratio only depends on the temperature. However, the relationship between the ratios and temperatures needs to be calibrated in advance; complicated by the existence of line shift and line broaden, these peaks need to be manually picked and recovered from the disturbance of line shape function $\phi_v$. 

In the following sections, we will evaluate the capability of SVPEN in solving the problem of gas quantification from MAS. First, we will introduce the detailed architecture and configurations of the SVPEN framework used in this application. Next, we will shift our attention in demonstrating the performance of SVPEN in inverse function and optimization mode.
\subsubsection{SVPEN configuration}\label{3.1.1}
\begin{figure}[hbt!]
\centering
\includegraphics[width=.7\textwidth]{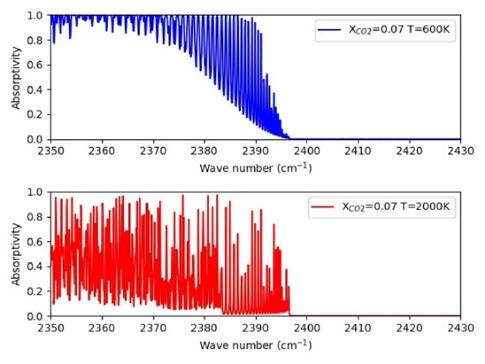}
\caption{ CO$_2$ absorption spectra at 600 K (top) and 2000 K (bottom), with a mole fraction of 0.07.}
\label{fig9}
\end{figure}

\textbf{Dataset for pretraining $G_1$}. The well-known HITEMP database ~\cite{rothmanHITEMPHightemperatureMolecular2010} and its paired simulation platform HAPI ~\cite{kochanovHITRANApplicationProgramming2016} were utilized to generate the spectra dataset for the pretraining $G_1$. The molecule chosen for this investigation is carbon dioxide (CO$_2$). The states, i.e., temperature and mole fraction, were assigned randomly in the range of 600-2000 K, and 0.05-0.07, respectively. The waveband selected is 2375-2395 $cm^{-1}$, since it is sensitive to temperature changes of CO$_2$ (Fig. ~\ref{fig9}). By setting an interval of 0.1 cm-1, the generated spectrum has 200 dimensions, and in total 10,000 samples were acquired.

\textbf{Network configurations}. Prior to constructing networks, a necessary step is to ascertain the formats of inputs and outputs. In this application, we normalized temperature, mole fraction and spectra to the range of 0-1. The reference boundaries to this normalization were respectively, the preconfigured 600-2000 K, 0.05-0.07, and the boundary values of the absorptivity $\alpha_v$ in the dataset generated. These normalized temperature ranges, mole fraction ranges, and spectra were then used as inputs and targets for the networks.

\begin{figure}[hbt!]
\centering
\includegraphics[width=1.0\textwidth]{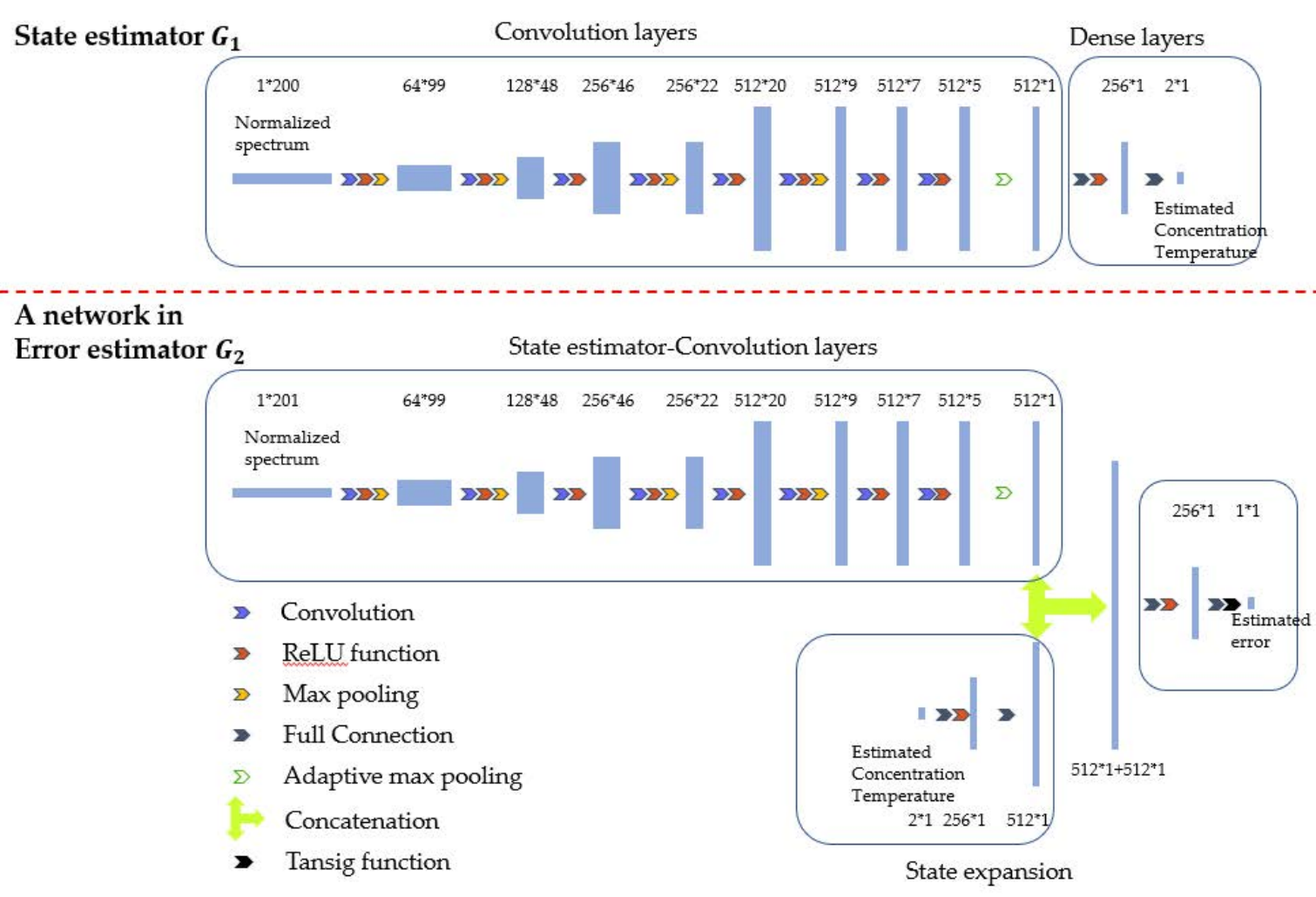}
\caption{ The architectures of networks for the solution of the MAS inverse problem.}
\label{fig10}
\end{figure}

The job of state estimator $G_1$ is to realize the mapping from the space of observations to the space of state estimates. Accordingly, $G_1$ takes the normalized spectrum $\alpha_{norm}\in R^{200}$ as the input and outputs the estimated normalized state $\hat{x}_{norm} \in R^2$, whose two entries were respectively the estimated temperature and mole fraction. Such a transformation was realized by using a classical convolutional neural network, VGG13 ~\cite{simonyanVeryDeepConvolutional2015}, as shown in Fig.~\ref{fig10}, where batch normalization was removed, 2D convolution layers were changed to 1D ones, and adaptive average pooling was added at the end of the convolution layers.

The error estimator $G_2$ has the functionality of taking the normalized spectrum and the estimated vector of normalized temperature and mole fraction as inputs and outputting the estimated error $e \in R^1$. The corresponding architecture of a single network inside is shown in Fig.~\ref{fig10}, where a branch-merge style is employed. As for the input branch of the normalized spectrum, the convolution layers of the state estimator $G_1$ are adopted, which transforms the normalized spectrum into a latent feature vector with a dimensionality of 512; as for the branch of estimated normalized state, it utilizes two convolution layers to embed $\hat{x}_norm$ into a latent feature vector with a dimensionality of 512. Next, both latent features are concatenated and transformed to the estimated error by the combination of the convolution and full connection layers. 

Overall, the architectures of both estimators are simple. Indeed, SVPEN can accommodate more complex architectures, such as transformers ~\cite{dosovitskiyImageWorth16x162021}, as per the requirements of the application. In this contribution, we utilize simple architectures, so as to emphasize the capacities of the framework instead of complex architecture of networks.

\textbf{Physical forward model.} Instead of HAPI, we embedded Radis ~\cite{pannierRADISNonequilibriumLinebyline2019}, another physical forward model, inside SVPEN. The decision for such a configuration is based on (a) Radis is more computationally efficient than HAPI, which favours optimization mode. (b) In pretraining dataset generation, to accelerate the computation of HAPI, the line intensities of HITEMP weaker than $1e^{-26}$  ($cm^{-1}/(molecule*cm^{-2})$) at the reference temperature 296 K were ignored. However, the high computational efficiency of Radis allows the use of all line intensities listed in HITEMP, therefore, more precise spectra can be simulated. Moreover, there are slight differences in the details of simulating line broaden, instrumental function, etc. between HAPI and Radis. Accordingly, the simulated spectra from the two platforms will be slightly different. This difference in turn can be utilized to demonstrate the reconfigurability of SVPEN by regarding Radis as an updated/alternative model of HAPI.

We recall that the estimated states from the state estimators are normalized, therefore, before feeding them to the physical forward model to estimate/reconstruct the observations, these need to be rescaled.

\textbf{Error calculation component.} As aforementioned, the function used to calculate the discrepancy between the estimated and original absorption spectra is a non-negative and monotonic function of the Euclidean distance (Eq.~\ref{eq24}, where $d_{\alpha}$ is the dimension of spectrum).The use of this function is inspired by the concept of potential wells in quantum mechanics, since the closer the reconstructed observation is to the actual observation, the more rapidly the error is reduced, which finally traps the estimation. By using such a trap-fashion metric, the convergence of the optimization mode is improved. As for the potential issue of getting trapped into local minima, the utilization of Cosine Annealing warmup start strategy can help.
\begin{equation}
\label{eq24}
    e_y=log(\frac{\|\alpha_{norm}-\hat{\alpha}_{norm}\|_2}{d_{\alpha}}+1)
\end{equation}

The only regularization utilized is the feasible domains of temperature and mole fraction. It is straightforward to practically identify the minima and maxima of these two parameters, and thus, the two regularization terms are consequently set as elaborated in Eqs. ~\ref{eq25}, ~\ref{eq26}.
\begin{equation}
\label{eq25}
    e_{reg,T}=Max(\hat{T}_{norm}-T_{upper,norm},0)+Max(T_{lower,norm}-\hat{T}_{norm},0)
\end{equation}
\begin{equation}
\label{eq26}
    e_{reg,X}=Max(\hat{X}_{norm}-X_{upper,norm},0)+Max(X_{lower,norm}-\hat{X}_{norm},0)
\end{equation}
\begin{equation}
\label{eq27}
    T_{upper, norm}=\frac{(T_{upper}-T_{ref,lower})}{(T_{ref,upper}-T_{ref,lower})}
\end{equation}

\begin{figure}[hbt!]
\centering
\includegraphics[width=.5\textwidth]{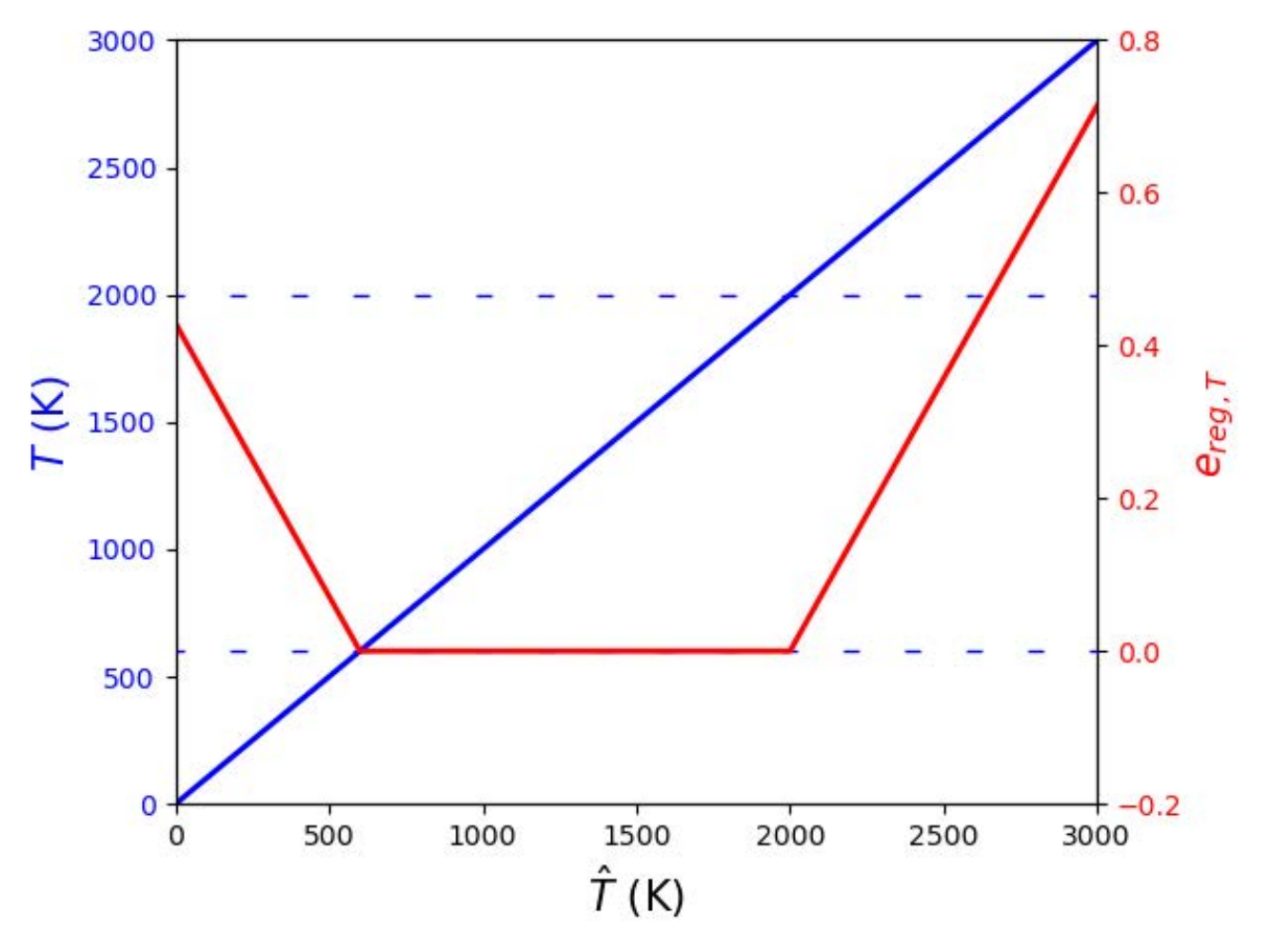}
\caption{ The use of feasible domain regularization}
\label{fig11}
\end{figure}

Where, $e_{reg,T}$  and $e_{reg,X}$ are the regularization constraints respectively in terms of the temperature and mole concentration. The subscripts $upper$ and $lower$ represent the feasible upper and lower boundaries for a state parameter. Since the estimated state from $G_1$  is normalized, the upper and lower boundaries of feasible range should also be normalized, an example is shown in Eq.~\ref{eq27}. Similar normalization can be done for the remaining feasible domain boundaries. The rationale of using such constraints is demonstrated in Fig.~\ref{fig11}. Suppose the feasible domain of the Temperature parameter is 600-2000 K, as shown by the range constrained by dash blue lines, thus, once the estimated temperature (blue line) is out of the range, the error term relating to the regularization parameter $e_{reg,T}$ (red line) increases, which in turn forces $G_1$ to iterate the estimation back into range. Such regularizations are useful for SVPEN to avoid convergence to invalid solutions and also reduce unnecessary search in irrelevant parts of the solution space.

The final error is a sum of the discrepancy and regularization terms (Eq.~\ref{eq28}). As the terms in Eqs.~\ref{eq24}-~\ref{eq26} are all normalized, the weights of each of the error terms can be simply set to one. Although the weights for regularization terms are usually relatively small and the sum of all weights is usually set to be one, we take this "rough" summation in order to construct a “rigid” boundaries of feasible domain, to assure the final converged solution is definitely inside.
\begin{equation}
\label{eq28}
    e=e_y+e_{reg,T}+e_{reg,X}
\end{equation}

\subsubsection{SVPEN performance}\label{3.1.2}
As already explained, SVPEN operates in two interconnected modes, i.e., inverse function and optimization. In this section, we first demonstrate the efficiency and accuracy of the inverse function mode by testing samples from the Independently and Identical Distribution (I.I.D) of the pretraining dataset. Moreover, we elaborate on the role of the physical forward model by explaining the physical reasonability of state estimations, which determines the shift between two modes. Following this, we test five progressively harder application scenarios outside the capability of inverse mode, to demonstrate the reconfigurability and capability of optimization mode. These five scenarios are respectively as follows: 
\begin{itemize}
\item (i) The MAS is mismatched with the physical model (labelled as mismatch in what follows) 
\item (ii) The MAS is from an outlier state but the same band (labelled as outlier in what follows)
\item (iii) The MAS is from an outlier state and another band (labelled as shifting band in what follows)
\item (iv) The MAS is from an outlier state and another larger band (labelled as expanding band in what follows)
\item (v) Using emission spectrum rather than MAS, and emission spectrum is from an outlier state and another larger band (labelled as emission spectrum in what follows).
\end{itemize}

\begin{figure}[hbt!]
\centering
\includegraphics[width=.6\textwidth]{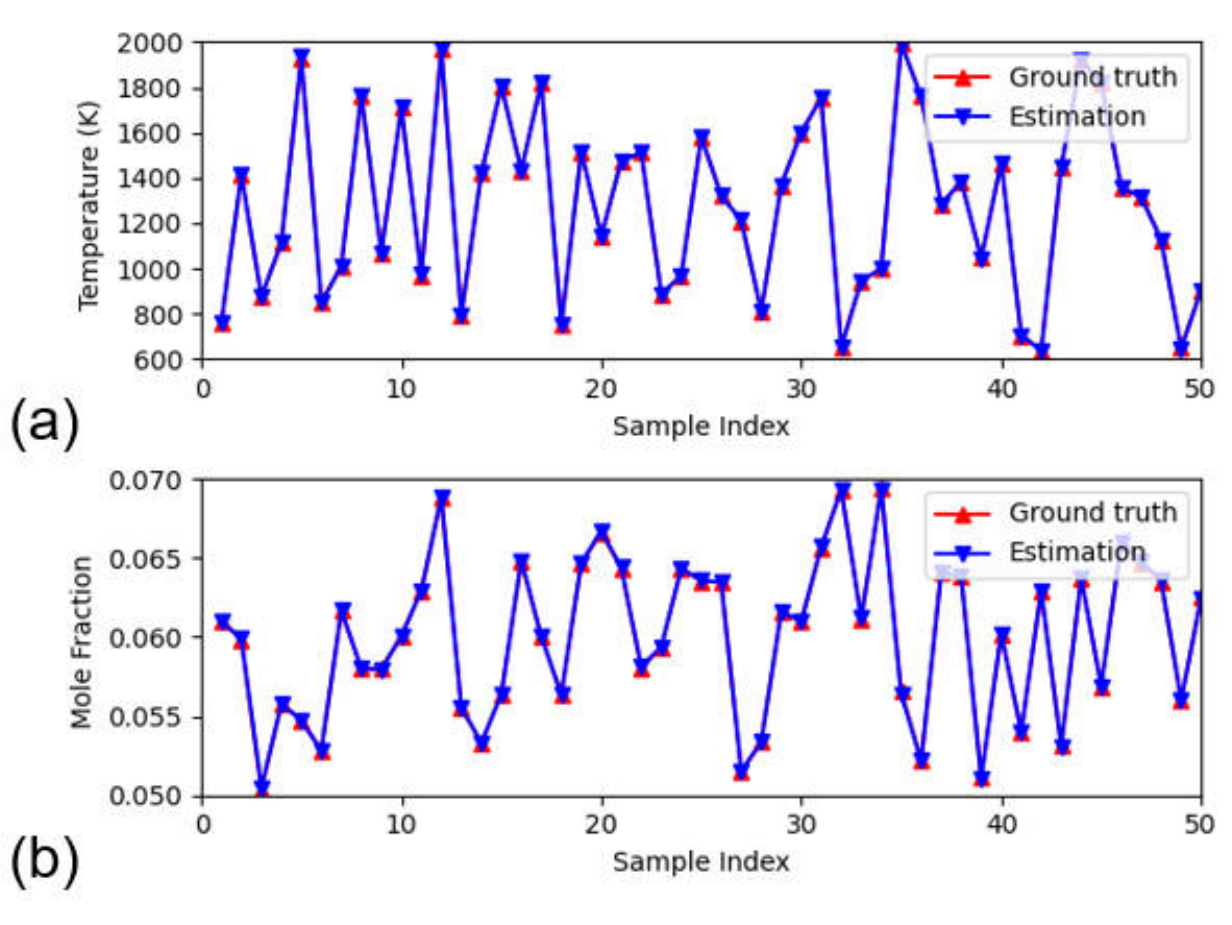}
\caption{ State estimation provided by the inverse function mode. (a) estimated temperatures and corresponding ground truth; (b) estimated mole fractions and associated ground truths.}
\label{fig12}
\end{figure}

\textbf{Inverse function mode.} 10,000 generated samples were divided into the training, validation, and test set with the ratio of 70\%,15\%, and 15\%, respectively. We first pretrained $G_1$ with the training dataset. Next, test set, which obviously satisfies I.I.D of training set, is utilized to test the inverse function mode of SVPEN. As $G_1$ is well-trained, SVPEN can provide accurate estimations of both temperature and mole fraction. As shown in Fig.~\ref{fig12}, fifty randomly picked test samples and their estimations are visually overlapped.

\begin{figure}[hbt!]
\centering
\includegraphics[width=.9\textwidth]{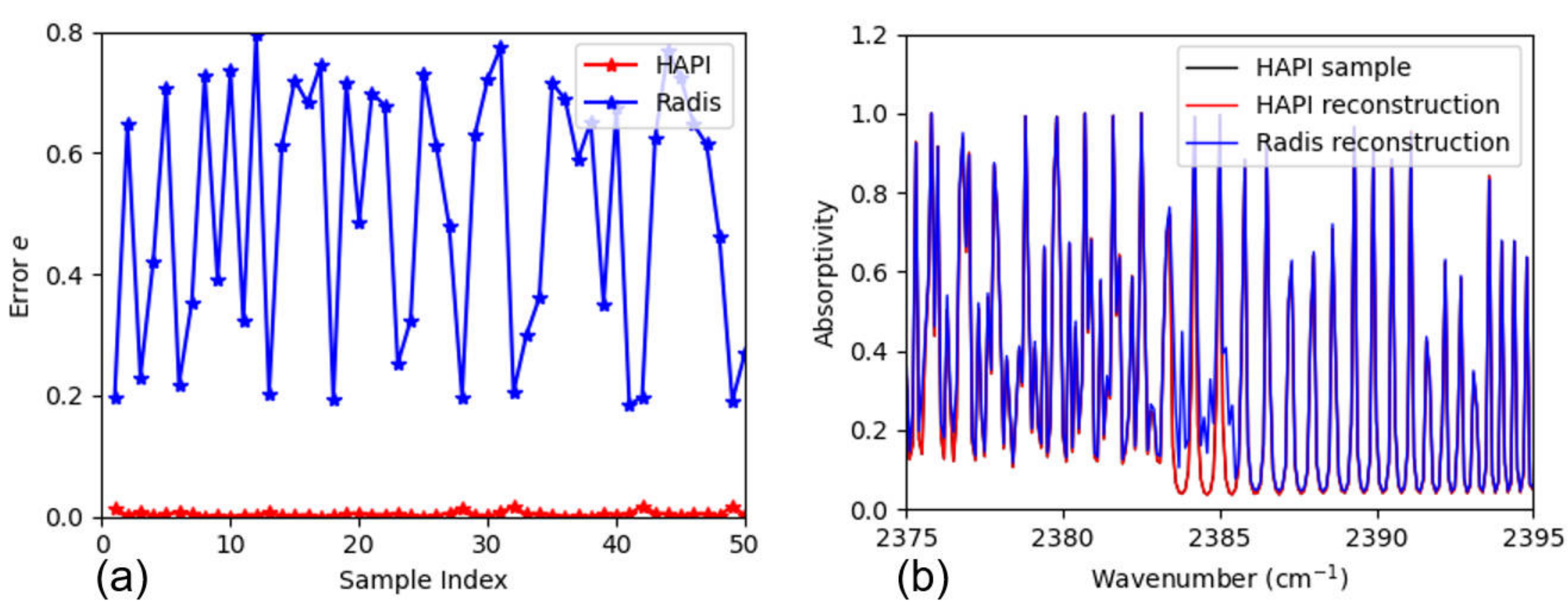}
\caption{ The estimation errors due to the use of different physical forward models.}
\label{fig13}
\end{figure}

However, the calculated errors are not directly determined by these state estimations, since no ground truths of states are available in actual test scenarios; instead, the calculated errors are mainly determined from the discrepancies between the given and estimated (reconstructed) observations, where the estimated observations are provided through the use of the physical forward model. Therefore, affected by the transformation/simulation differences between different physical models, e.g., HAPI and Radis, the same state estimations provided by $G_1$ can lead to distinct errors. For example, as shown in Fig.~\ref{fig13} (a), for the estimations of these fifty test samples, the calculated errors  from using HAPI as the forward model are no more than 0.02, however, when replacing HAPI by Radis, the calculated errors are substantially increased. When picking one concrete test sample out, such as a spectrum generated by HAPI with the state of 1500 K and a mole fraction of 0.06, the corresponding estimation from $G_1$ is 1500.3 K and 0.06. As shown in Fig.~\ref{fig13} (b), the reconstructed spectrum from HAPI is visually overlapping with the test sample, the calculated error is 0.0015, whereas the one reconstructed from Radis has apparent difference to the test sample, the calculated error is 0.685, much larger than the error of 0.0015. This example highlights the importance of the physical forward model in the context of the inverse function mode in assessing the quality of state estimation. 

\textbf{(i) Case of Mismatch}. In this case, we focus on the error caused by the mismatch between the test sample and the reconstructed one from the physical forward modelling. In fact, this mismatch is common and somewhat expected, since it is almost impossible for a theoretical model/simulation process to accurately consider all factors behind an actual physical phenomenon, e.g., a variety of errors exist when collecting experimental observations. These mismatches between the test data and the simulated data can be encapsulated under the term of uncertainties ~\cite{ghahramaniProbabilisticMachineLearning2015}. 

The existence of uncertainties, in turn, emphasizes the necessity of the optimization mode, since for a new set of observations collected during an experiment, we cannot assure they satisfy the I.I.D assumption of training set of $G_1$, especially with uncertainties, and usually, we do not know the ground truth of the corresponding states. Rather than believing $G_1$ is a precise, intelligent and authoritative predictor which can tackle all cases, we have to admit that, in most of cases, $G_1$ is blinded by uncertainties; instead, the thing we human would more like to believe is physics, which is the tool we explain the world. This tool herein is represented by, an estimated state can be regarded to be acceptable, under the condition that the corresponding reconstructed observation through the process of physical forward model is similar to the actual observation. This acceptable solution is usually done by iterations under the scope of optimization mode, since $G_1$ is almost blind under the attack of uncertainties.  In the language of machine learning, inverse function mode works in an unsupervised way although its training is supervised, but optimization mode always works in the way that supervised by the validation of physical models.

Accordingly, facing with the obvious error caused by the presence of uncertainties, SVPEN swiches from inverse function to optimization mode. It is interesting to observe that the minimum error after 1000 iteration is 0.578, and the corresponding estimated values of temperature and mole fraction are 1370.40 K and 0.0439, respectively ( Table~\ref{Table1}). Given that the ground truth of the state is 1500 K and 0.06, respectively, it seems that the optimized estimation is worse in accuracy than the first guess provided by the inverse function mode. However, this estimation is more physically reasonable than the guess provided by the inverse function mode, because it accommodates for model mismatch and leads to the closest approximation of a given observation that can be provided by Radis. The imperfection in state estimation is due to the “low modelling fidelity” of the physical forward model to the test spectrum, instead of an issue of the optimization mode.

\begin{table}
\caption{ Inverse modelling results for the various cases}
\label{Table1}
\centering
\resizebox{\textwidth}{!}{\begin{tabular}{ccccccc} 
\hline
Test sample              & HAPI                                                          & \multicolumn{5}{c}{Radis}                                                                                                                                                                                                                                                                                                    \\ 
\hline
Types                    & (i) Mismatching                                               & \multicolumn{2}{c}{(ii)~ Outlier}                                                                                            & (iii) Shifting band                                           & (iv) Expanding band                                           & (v)~ Emission spectrum                                        \\
Ground truth             & 1500, 0.06                                                    & 300,0.03                                                      & 3000,0.3                                                     & 300,0.03                                                      & 300,0.03                                                      & 3000,0.3                                                      \\
waveband                 & 2375-2395                                                     & 2375-2395                                                     & 2375-2395                                                    & 2330-2350                                                     & 2330-2370                                                     & 2330-2370                                                     \\
Acceptable accuracy      & 0.15                                                          & 0.15                                                          & 0.1                                                          & 0.15                                                          & 0.15                                                          & 0.15                                                          \\
Feasible domain          & \begin{tabular}[c]{@{}c@{}}300-2000 K\\0.01-0.15\end{tabular} & \begin{tabular}[c]{@{}c@{}}100-1000 K\\~0.01-0.1\end{tabular} & \begin{tabular}[c]{@{}c@{}}2000-4000K \\0.1-0.4\end{tabular} & \begin{tabular}[c]{@{}c@{}}100-1000 K \\0.01-0.1\end{tabular} & \begin{tabular}[c]{@{}c@{}}100-1000 K \\0.01-0.1\end{tabular} & \begin{tabular}[c]{@{}c@{}}2000-4000K \\0.1-0.4\end{tabular}  \\
First state
  estimation & 1500.3K 0.06                                                  & 1855.9 K 0.075                                                & 1851.6 K 0.09                                                & 987.8 K 0.125                                                 & 4799.6 K 0.4                                                  & 4799.6 K 0.382                                                \\
Final state
  estimation & 1370.4 K 0.044                                                & 296.1 K 0.031                                                 & 3042.4 K 0.328                                               & 310.9 K 0.033                                                 & 283.1 K 0.028                                                 & 2976.7 K 0.304                                                \\
Final error              & 0.578                                                         & 0.141                                                         & 0.099                                                        & 0.14                                                          & 0.1                                                           & 0.135                                                         \\
Iterations               & 971/1000                                                      & 588                                                           & 111                                                          & 245                                                           & 441                                                           & 602                                                           \\
\hline
\end{tabular}}
\end{table}

\begin{figure}[hbt!]
\centering
\includegraphics[width=.6\textwidth]{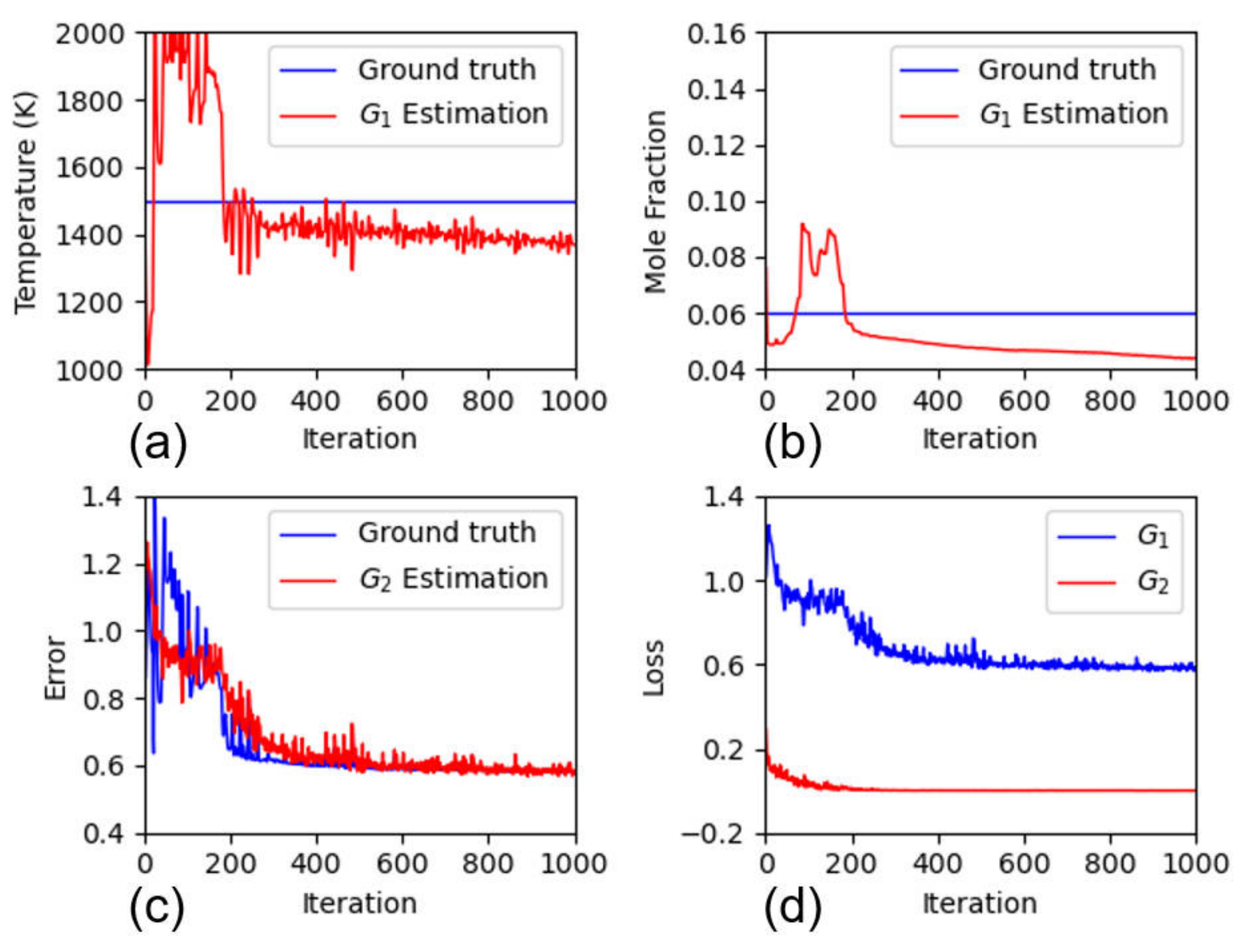}
\caption{ Detailed optimization process for the case of mismatch (a) the iteration process of temperature estimation; (b) the iteration process of mole fraction; (c) the changes of errors in iteration process; (d) the changes of loss functions of $G_1$ and $G_2$ during iterations}
\label{fig14}
\end{figure}

The detailed iteration process of this mismatch case is demonstrated in Fig.~\ref{fig14}, where the changes in estimated temperature, estimated mole fraction, the actual and estimated errors, and the losses of both estimators are shown in (a), (b), (c), and (d), respectively. It can be seen that although the estimations of temperature and mole fraction are getting farther away from the ground truth, the error ground truth provided by the physical comparison module is gradually decreasing (Fig. ~\ref{fig14} (c)). Along with that, the estimated error is also decreasing, which proves that the state estimator $G_1$ is devoted to decreasing the estimated error, as shown in Fig.~\ref{fig14} (c) and blue line of Fig.~\ref{fig14} (d). In addition, the error estimator $G_2$ gradually learns the means to estimate the error, which is indicated by the fact that the estimated error starts to gradually overlap with the actual error visually, as shown in Fig.~\ref{fig14} (c), and the $Loss_{G_2}$, represented by the red line of Fig.~\ref{fig14} (d), decreases during the course of descent optimization.

\textbf{(ii) Presence of outliers.} In order to further demonstrate the characteristics and capability of SVPEN, it was applied to progressively sophisticated cases. The first case is estimating temperature and mole fraction from two MAS, generated by Radis with states of (300K, 0.03) and (3000 K, 0.3), respectively. Both the mole fraction and temperature are outside the reference ranges used to synthesize spectra for the pretraining of $G_1$. As shown in  Table~\ref{Table1}, the feasible domains for the purposes of imposing regularization are selected as 100-1000 K, 0.01-0.1, and 2000-4000 K, 0.1-0.4, respectively. According to the nonlinear mapping between error and state estimation discrepancies, setting error threshold is somehow empirical, which was set as 0.15 herein after multiple experiments on various cases, since this error threshold can lead to the relative error of state estimations no more than 10\% in most cases. In some cases, it can be further restricted, e.g., this error threshold was further decreased to 0.1 for the case of (3000 K, 0.3), since the value of 0.15 was found to be easily satisfied with very few iterations. In addition,one can also set an unachievable error threshold, and then pick the optimal one in iterations as the acceptable one, as what is done in the case of mismatch and section ~\ref{3.2}. 

\begin{figure}[hbt!]
\centering
\includegraphics[width=.6\textwidth]{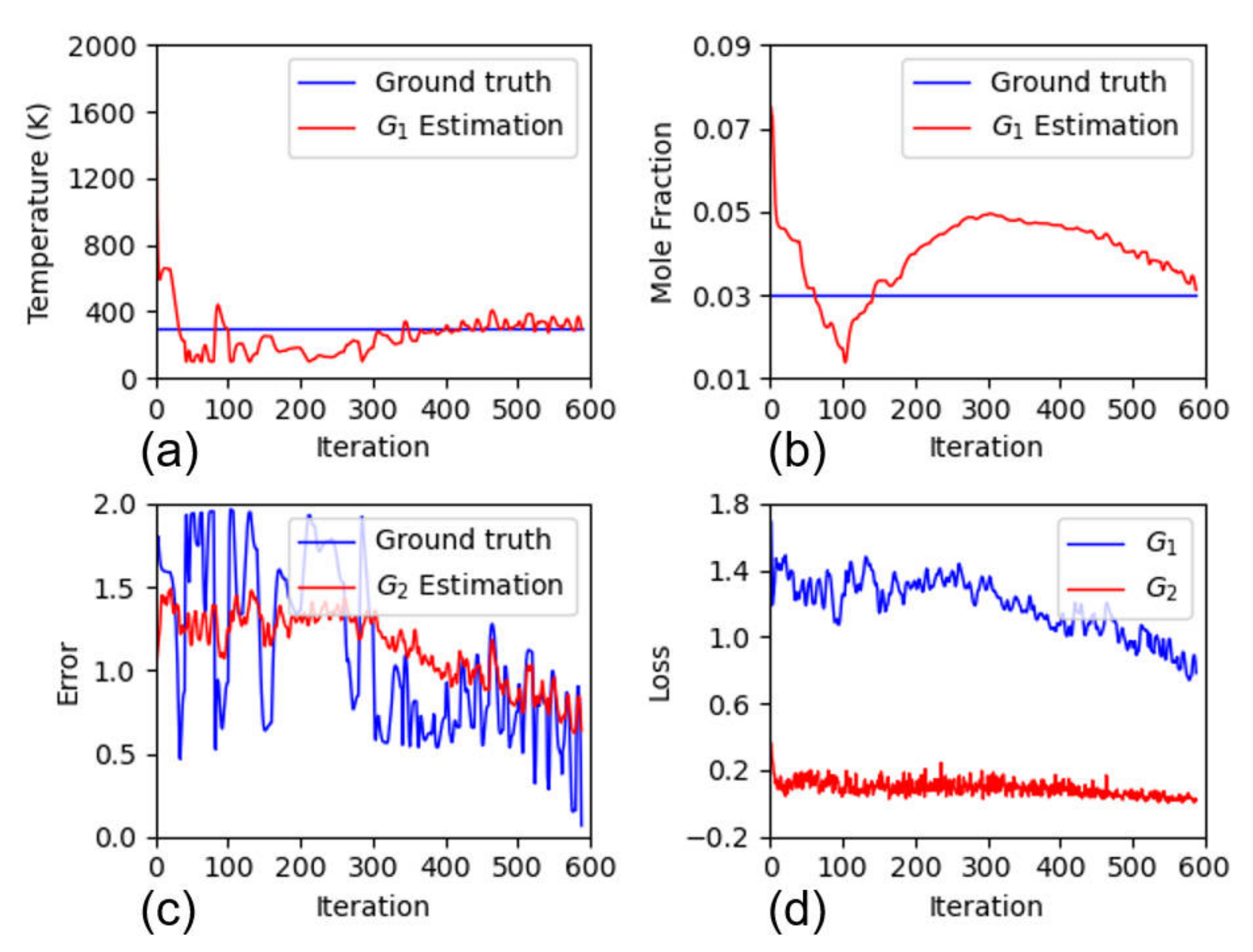}
\caption{ Detailed optimization process for the presence of outlier: MAS generated from 300K, 0.03. (a) the iteration process of temperature estimation; (b) the iteration process of mole fraction; (c) the changes of errors in iteration process; (d) the changes of loss functions of $G_1$ and $G_2$ during iterations}
\label{fig15}
\end{figure}

The initial estimation given by the inverse function mode were surprisingly blind. The predictions in the two cases were (1855.9 K, 0.075) (Fig. ~\ref{fig15}), and (1851.6 K, 0.090), respectively. These results indicate that $G_1$ is not good at handling outliers. Indeed, this is where the optimization mode comes in to guide $G_1$ towards the right estimation direction. As shown in Fig.~\ref{fig15}, for the case of MAS from (300 K, 0.03), $G_2$ gradually learns the way to estimate the error (Fig. ~\ref{fig15} (c)), although not exactly the same as the actual one from physical evaluation module (ground truth in Fig.~\ref{fig15} (c)) , it gives $G_1$ the feedback on the state estimation accuracy, so leads the estimated state eventually converges to (296.1 K, 0.031), with the corresponding error of 0.141; for the MAS from (3000 K, 0.3), the estimated state eventually converges to(3042.4 K, 0.328), with corresponding errors of 0.099 ( Table~\ref{Table1}).

\begin{figure}[hbt!]
\centering
\includegraphics[width=.6\textwidth]{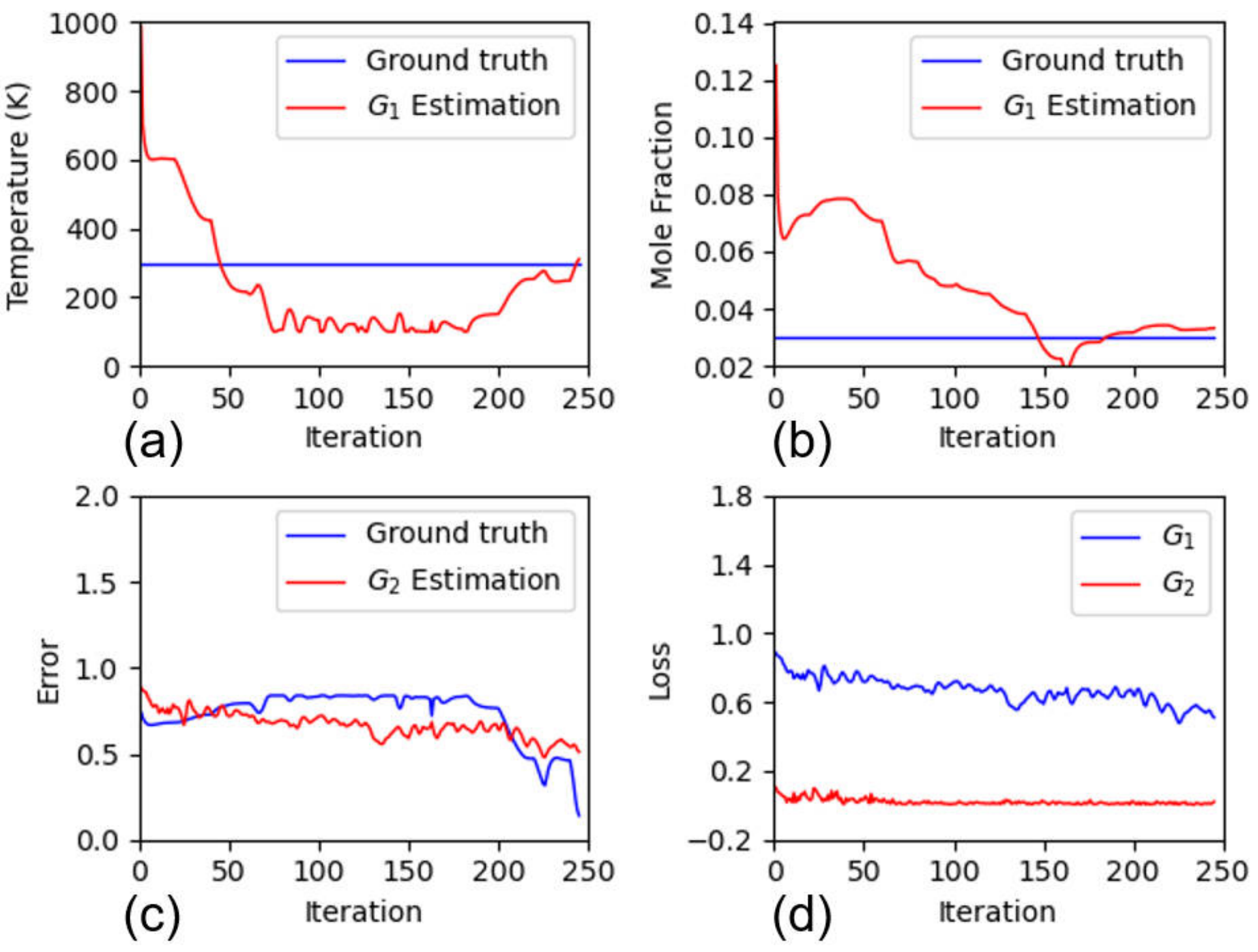}
\caption{ Detailed optimization process for the case of shifting bands. (a) the iteration process of temperature estimation; (b) the iteration process of mole fraction; (c) the changes of errors in iteration process; (d) the changes of loss functions of $G_1$ and $G_2$ during iterations}
\label{fig16}
\end{figure}

\textbf{(iii) Shifting bands.} The test spectrum generated by (300K, 0.03) was further shifted to another waveband, i.e., 2330-2350 $cm^{-1}$. Correspondingly, the waveband set in the physical forward model was also changed to 2330-2350 $cm^{-1}$. As a consequence, any state estimated by $G_1$ in this case would be baseless and meaningless since the physical model utilized is completely different from the one used for the pretraining of $G_1$. Therefore, as shown in  Table~\ref{Table1}, the initial estimation of $G_1$ is (987.8 K, 0.125), but with the support of optimization mode, the estimated state converges to (310.9 K, 0.033) after 245 iterations (Fig. ~\ref{fig16}), with an error of 0.140.

\begin{figure}[hbt!]
\centering
\includegraphics[width=.6\textwidth]{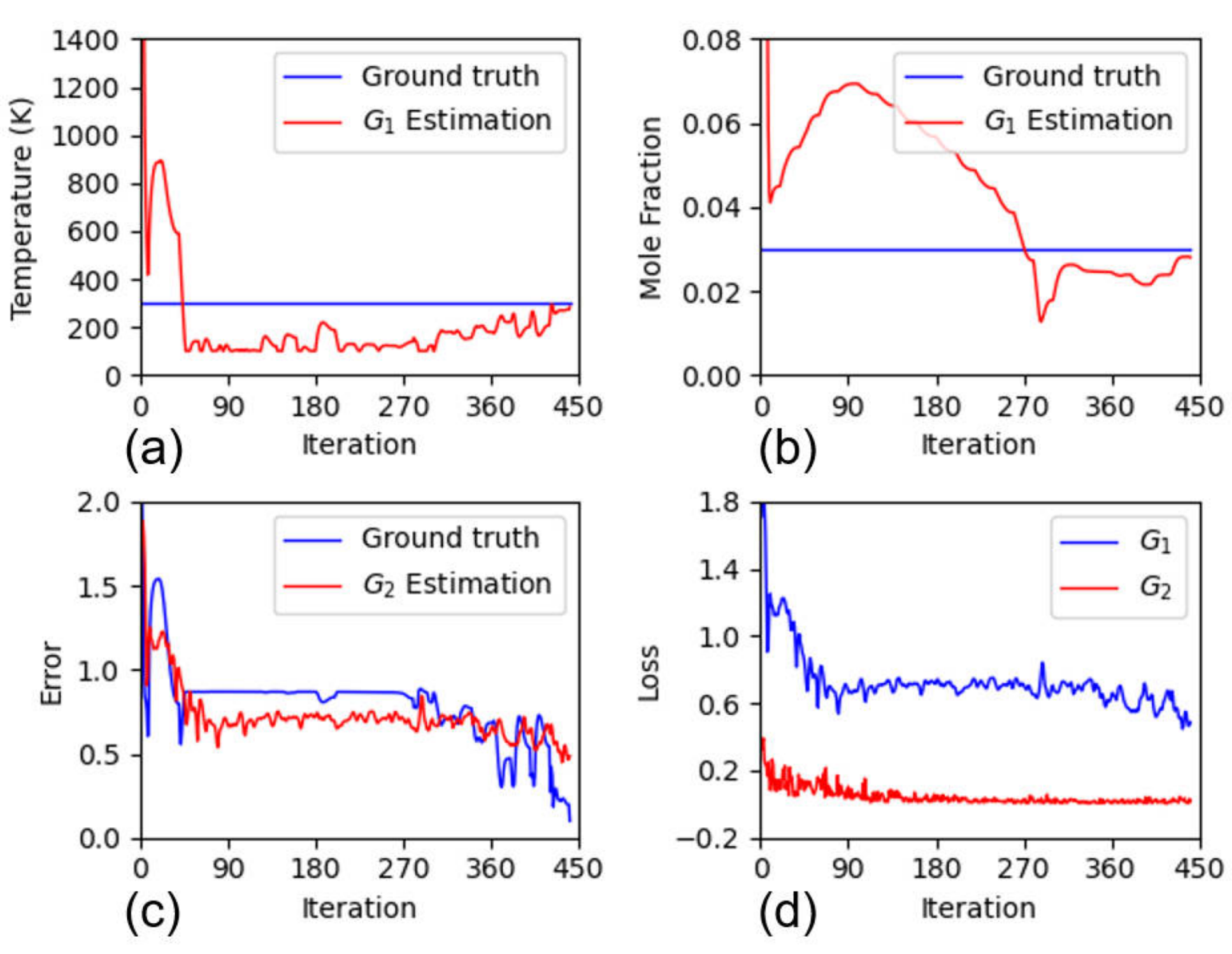}
\caption{ Details of the optimization process in the case of expanding wavebands. (a) the iteration process of temperature estimation; (b) the iteration process of mole fraction; (c) error changes as a function of the number of iterations; (d) the changes of loss functions of $G_1$ and $G_2$ during iterations.}
\label{fig17}
\end{figure}

\textbf{(iv) Expanding bands.} The waveband of the test spectrum was doubled by expanding from 2330-2350 $cm^{-1}$ to 2330-2370 $cm^{-1}$. As adaptive average pooling was used to construct the networks (see Fig.~\ref{fig10}), so that SVPEN is expected to handle various input dimension, however, as we all known, the $G_1$ cannot give reasonable estimations as its trained with an preset input dimension. Therefore, in this case, the first estimation provided by $G_1$ is (4799.6 K, 0.40), unsurprisingly outrageous values compared to ground truth of (300 K, 0.03) (Table~\ref{Table1}). Accordingly, SVPEN switches to optimization mode and finally converges to (283.1 K, 0.028) in 441 iterations. Fig.~\ref{fig17} elaborates the details of the iteration process. We can tell that SVPEN is able to correct the extremely biased initial estimates to pretty accurate estimates of temperature and mole fraction through cooperation of $G_1$ and $G_2$. As demonstrating in other cases, $G_2$ does not need to provide very accurate error estimations, instead, merely need to sense the magnitude and approximated change of the error. This suffices in guiding $G_1$ to accurately estimate temperature and mole fraction.

\textbf{(v) Using emission spectra.} In this case study, the absorption spectrum was replaced by an emission spectrum, another spectroscopy technique. Based on the physical forward model of MAS, a simple forward model of emission spectroscopy can be formulated by adding two more equations ~\cite{kangEmissionQuantificationPassive2022}, i.e., Eqs.~\ref{eq29}~\ref{eq30}.
\begin{equation}
\label{eq29}
    I_{E,v}=I_{B,v}(1-\alpha_v)
\end{equation}
\begin{equation}
\label{eq30}
    I_{B,v}=\frac{2hv^3}{c^2[exp(\frac{hv}{k_BT})-1]}
\end{equation}

where $I_E$ and $I_B$ are the emission intensity and blackbody radiation, respectively. By comparing the forward models of emission and absorption spectroscopy, it is evident that emission spectroscopy involves a higher degree of nonlinearity with respect to the parameters of mole fraction and temperature. Besides, the emission intensity is an unconstrained absolute value, whereas the absorptivity in cases above is a relative value which is in the range of 0-1. Both of these observations make retrieving temperature and mole fraction from emission spectroscopy more difficult. 

\begin{figure}[hbt!]
\centering
\includegraphics[width=.6\textwidth]{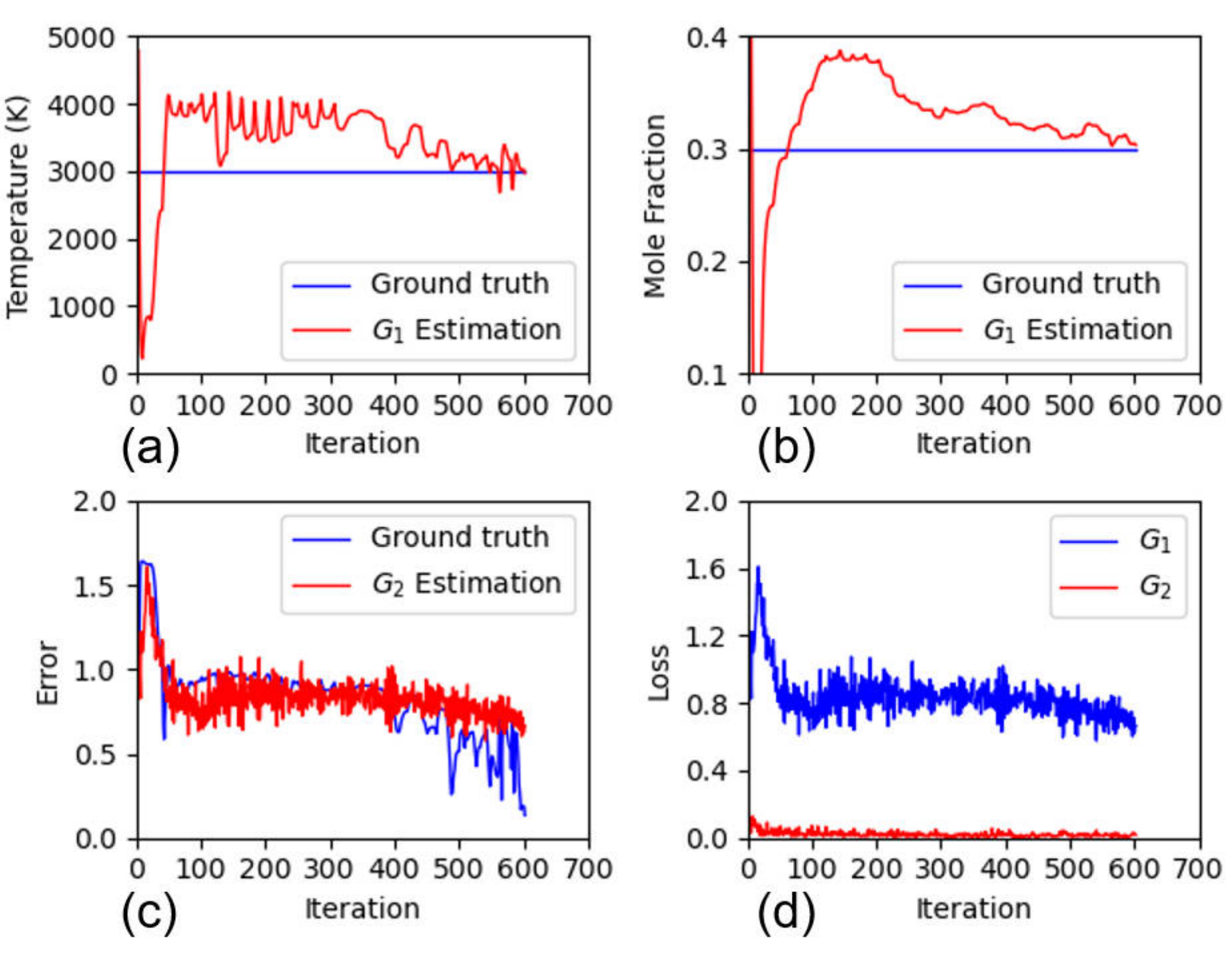}
\caption{ Details of the optimization process for the case of emission spectrum. (a) estimation performance of temperature estimation as a function of iterations; (b) estimation performance of mole fraction as a function of iterations; (c) error changes as a function of iterations; (d) the changes of loss functions of $G_1$ and $G_2$ during iterations}
\label{fig18}
\end{figure}

The test emission spectrum was still synthesized in the doubled waveband of 2330-2370 $cm^{-1}$, and the state used was shifted to (3000 K, 0.3) instead of (300 K, 0.03), as temperature of 300 K cannot provide sufficient blackbody radiation in the waveband of 2330-2370 $cm^{-1}$. By examining  Table~\ref{Table1}, as expected, the initial estimate provided in the inverse function mode is rather biased, i.e., (4799.6 K, 0.382). Albeit emission spectroscopy is substantially different to absorption spectroscopy, when SVPEN switches to optimization mode, it is able to converge to a state estimation of (2976.7 K, 0.304) after 602 iterations. As shown in Fig.~\ref{fig18}, the initial estimation of the inverse function mode causes a huge error, which in turn leads to greedy correction of the estimated state, thus overcompensating the estimation of temperature and mole fraction. Proceeding with that, SVPEN gradually discovers the route to provide physically reasonable and fairly accurate estimations of temperature and mole fraction.

\begin{figure}[hbt!]
\centering
\includegraphics[width=.7\textwidth]{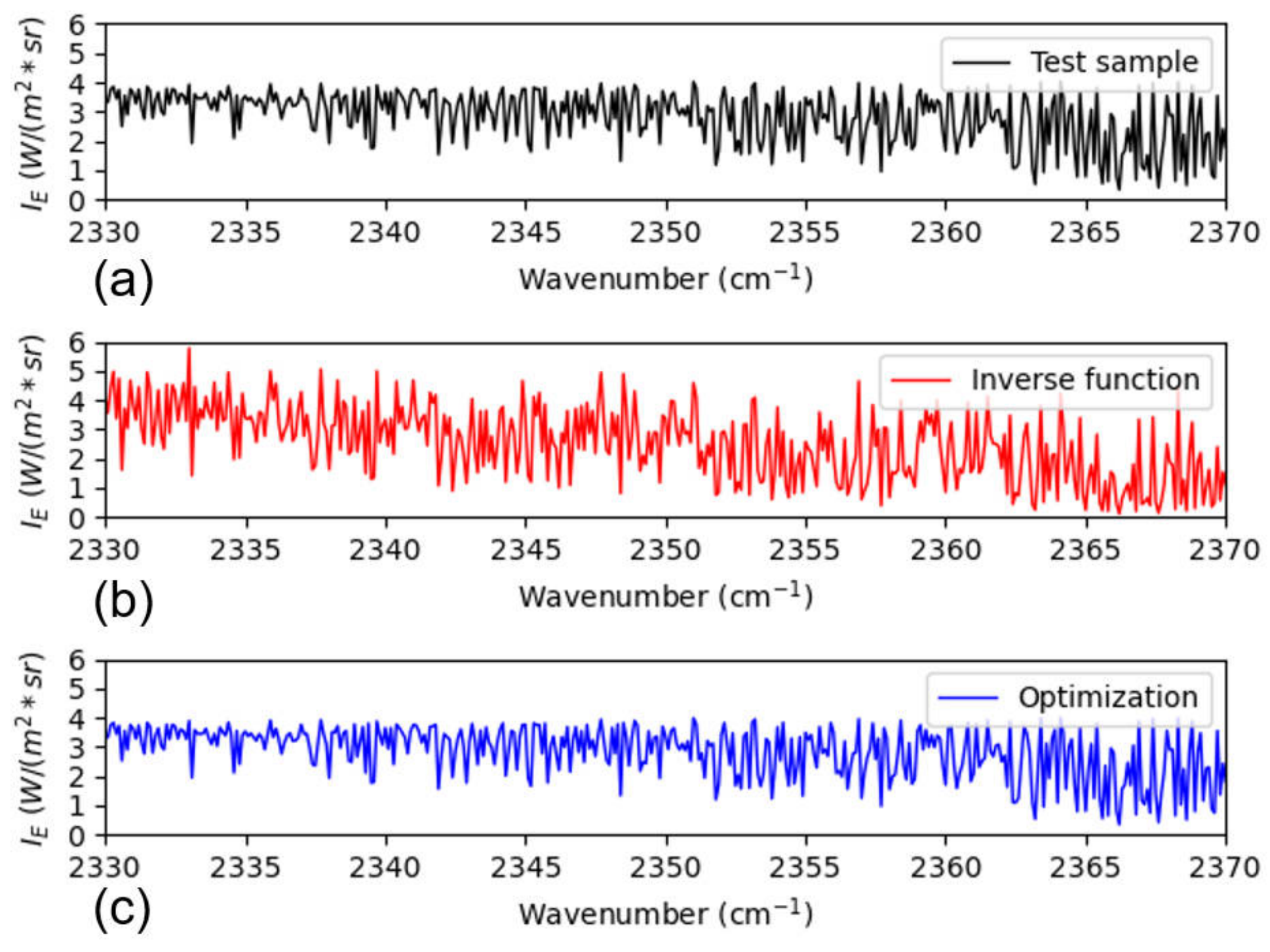}
\caption{ The emission spectra reconstructed from the state estimates by the inverse function mode and the optimization mode. (a)Test spectrum for a given observation (b) reconstructed spectrum by Radis using the estimates of (4799.6 K, 0.382), provided by the inverse function mode (c) reconstructed spectrum by Radis using the estimates of (2976.7 K, 0.304), provided by the optimization mode.}
\label{fig19}
\end{figure}

Fig. ~\ref{fig19} presents the ground truth of the emission spectrum, and its reconstructed result using the estimations of inverse function and optimization modes. From the comparison between these plots, one could appreciate the need for and capabilities of the optimization mode according to the thorough differences between the groun truth and the one reconstructed from inverse function mode. Besides, comparing these emission spectra to the absorption spectra shown in Fig.~\ref{fig13}, one can intuitively understand how different the application of SVPEN to emission spectroscopy from its intended design is, which embodies the reconfigurability of SVPEN.

From the demonstrations of case studies above, one can sense the characteristics and capability of SVPEN: The inverse function mode of SVPEN offers the same advantages as supervised learning-based inverse modelling methods, i.e., it provides efficient estimates of the state, e.g., temperature and mole fraction. And the estimates can be sufficiently accurate when the test samples can be regarded as picked from I.I.D of pretraining set of $G_1$. Although such an I.I.D assumption cannot be always valid, the optimization mode of SVPEN will take the turn and assure to provide physically reasonable estimations of states through iterations. And also, this optimization mode provides the reconfigurability of SVPEN, as shown here, SVPEN can work well regardless of changes, e.g., using different simulation platforms, wavebands, and even spectroscopy types, without requiring any repeating of pretraining the state estimator $G_1$. It is notable that one of the key steps for realizing SVPEN is the self-validation of the physical evaluation module, which is not only the judge to decide whether a state estimation is physically reasonable to accept, but also the guider that leading the optimization direction of SVPEN.

\subsection{Application II: Turbofan cycle analysis}\label{3.2}
Turbofan ~\cite{linke-diesingerSystemsCommercialTurbofan2008}, one of most complex gas turbine engine systems, is the dominant propulsion system favored by commercial airliners. To simulate, analyze and design gas turbine engines, a gas turbine forward model is needed. The forward model is usually formulated through aerodynamic and thermodynamic modelling of the components in turbofan. Taking the two-spool turbofan: CFM56-7B, as an example, the components that need to be modelled include inlet, fan, low pressure and high pressure compressors, combustor, high pressure and low pressure turbines, core and fan nozzles. In addition, energy losses in all kinds of mechanical connections/ducts also need to be considered. It can be appreciated that the forward modelling of the turbofan is a sophisticated and nonlinear process, which can be abstractedly described as Eq.~\ref{eq31}.
\begin{equation}
\label{eq31}
    F(x_E,x_C,x_{\eta})=y_P
\end{equation}

Where, $x_E,x_C$  and $x_{\eta}$ represent the environmental, cycle, and component parameters, respectively. More specifically, $x_E$ mainly includes the flight altitude, temperature, Mach number, etc. $x_C$ primarily considers the pressure ratio of each component, and the maximum temperature reached by combustion. $x_{\eta}$ mainly includes the efficiency of every component inside the engine. $y_P$ is the performance of the engine, i.e., the observation herein, which is basically constituted of the thrust Force ($F$), Thrust Specific Fuel Consumption ($TSFC$), in advanced models, it may also includes metrics of noise and emission.

Cycle analysis ~\cite{mattinglyAircraftEngineDesign2018} of turbofan design is the corresponding inverse problem. In cycle analysis, the required performance $y_{P,req}$ is given, and a judiciously considered operation environment $x_E$ is assigned. In this case the task is to find the state, i.e., $x_C$ and $x_{\eta}$, to satisfy  $y_{P,req}$, which can be simply elaborated by Eq. ~\ref{eq32}.
\begin{equation}
\label{eq32}
    F(x_C,x_{\eta})=y_{P,req}|x_E
\end{equation}

Unlike the case of MAS, this inverse problem is rank-deficient and ill-posed, since a two-spool turbofan engine, e.g., CFM56, has eleven CAC parameters to ascertain, whereas the performance parameters may be two, i.e., $F$ and $TSFC$, as shown in  Table~\ref{Table2}. The nature of this problem indicates that infinite feasible solutions of Cycle and Component (CAC) parameters exist. In other words, cycle analysis has no absolutely correct answer. Instead, a variety of solutions may be admissible, since the final design is a comprehensive consideration of performance, technological and economic issues, design preferences, etc. The “soft” posing of the cycle analysis problem makes traditional inverse function methods inappropriate for its solution. 

The conventional method to solve this problem is through engineers’ semi-empirical trials with following procedures: (a) Reducing the search space. the adjustments can only be doen to limited CAC parameters with the reference to a baseline engine ~\cite{debiasiCycleAnalysisQuieter2001} and the component parameters is often preset or correlated to cycle parameters ~\cite{liewParametricCycleAnalysis2005,mattinglyAircraftEngineDesign2018}, both of these operations make the search space for a proper design highly constrained and reduced.  (b) Then performing optimization in such a reduced space. This involves doing sensitive analysis to understanding of the physical mechanism or trends by changing various parameters, and determining a group of CAC parameters to satisfy the performance requirements ~\cite{liewParametricCycleAnalysis2005,mattinglyAircraftEngineDesign2018,debiasiCycleAnalysisQuieter2001}. However, such an approach may not lead to the optimum, but result to an unbalanced match between parameters. For example, the CAC parameters of CFM56-7B provided in ~\cite{aydinExergeticSustainabilityIndicators2015} overestimate the efficiencies of turbines ~\cite{kumarInnovativeApproachesReducing2012}, but underestimate the combustion efficiency ~\cite{mongiaTAPSFourthGeneration2003} ( Table~\ref{Table2}).

\begin{table}
\centering
\caption{ CFM 56-7B CAC parameters estimation}
\label{Table2}
\begin{tabular}{cccc} 
\hline
Category                             & parameter & Meaning                                           & Estimation \textsuperscript{~\cite{aydinExergeticSustainabilityIndicators2015}}  \\ 
\hline
\multirow{5}{*}{Cycle parameter}     & $BPR$         & Bypass ratio                                      & 5.313                            \\
                                     & $\pi_{fan}$         & Pressure ratio of fan                             & 1.636                            \\
                                     & $\pi_{LC}$        & Pressure ratio of low-pressure compressor         & 2.84                             \\
                                     & $\pi_{HC}$         & Pressure ratio of High-pressure compressor        & 9                                \\
                                     & $T_{max}$         & Highest temperature in the burner                 & 1624 K                           \\
\multirow{6}{*}{Component parameter} & $\eta_{fan}$         & Isentropic efficiency of fan                      & 0.864                            \\
                                     & $\eta_{LC}$         & Isentropic efficiency of low-pressure compressor  & 0.87                             \\
                                     & $\eta_{HC}$         & Isentropic efficiency of high-pressure compressor & 0.915                            \\
                                     & $\eta_{B}$         & Isentropic efficiency of burner                   & 0.85                             \\
                                     & $\eta_{HT}$        & Isentropic efficiency of high-pressure turbine    & 0.985                            \\
                                     & $\eta_{LT}$        & Isentropic efficiency of low-pressure turbine     & 0.985                            \\
\multirow{4}{*}{Fixed parameter}     & $m_{total}$        & Total mass flow                                   & 361 kg/s                         \\
                                     & $\eta_{inlet}$        & Isentropic efficiency of inlet                    & 0.98*                            \\
                                     & $\eta_{CN}$       & Isentropic efficiency of core nozzle              & 0.985*                           \\
                                     & $\eta_{FN}$        & Isentropic efficiency of fan nozzle               & 0.99*                            \\
Expected Performance                 & \multicolumn{3}{c}{$F$: 121 kN; $TSFC$: 10.63 g/(kN.s) \textsuperscript{64}}                        \\
\multicolumn{4}{c}{*: Estimated values by the authors of current research}                                                              \\
\hline
\end{tabular}
\end{table}

In this section, we instead used SVPEN to perform cycle analysis. To some extent, this application is a perfect showcase to demonstrate the power of the optimization mode, since no ground truth of observations can be provided to pretrain $G_1$, and thus inverse function mode is thoroughly abandoned. In the following subsections, we first introduce the configuration of SVPEN for solving this problem, and then proceeding with demonstrating how to use the reconfigurability of SVPEN to satisfy various design constraints.

\subsubsection{SVPEN configuration}\label{3.2.1}
\textbf{Network configurations} The state estimator $G_1$ receives the observation, $F$ and $TSFC$, as the inputs, and provides the estimations of eleven CAC parameters as outputs. In practice, the input observation is self-normalized, as demonstrated in Eq.~\ref{eq33}.
\begin{equation}
\label{eq33}
    y_{P,req,norm}=\frac{y_{P,req}}{y_{P,req}}=1
\end{equation}

Thus, (1,1), which corresponding to the normalized values of $F$ and $TSFC$, is always used as the inputs for the $G_1$. Meanwhile, the estimated observations from physical model were also normalized, and then used to calculate errors. The output of $G_1$ is the estimated normalized CAC parameters $\hat{x}_{CAC,norm}$, where the actual estimated CAC parameter ( $\hat{x}_{CAC}$) were acquired by transforming the normalized values with the feasible domain of CAC parameters which will be defined in section ~\ref{3.2.2}, as shown in Eq.~\ref{eq34}.
\begin{equation}
\label{eq34}
    \hat{x}_{CAC}= \hat{x}_{CAC,norm}*(x_{CAC,upper}-x_{CAC,lower})+x_{CAC,lower}
\end{equation}

By using this definition, the upper and lower boundaries of the feasible domain of CAC parameters, i.e., $x_{CAC,upper}$  and $x_{CAC,lower}$, have the normalized values of 1 and 0, respectively.

\begin{figure}[hbt!]
\centering
\includegraphics[width=1.0\textwidth]{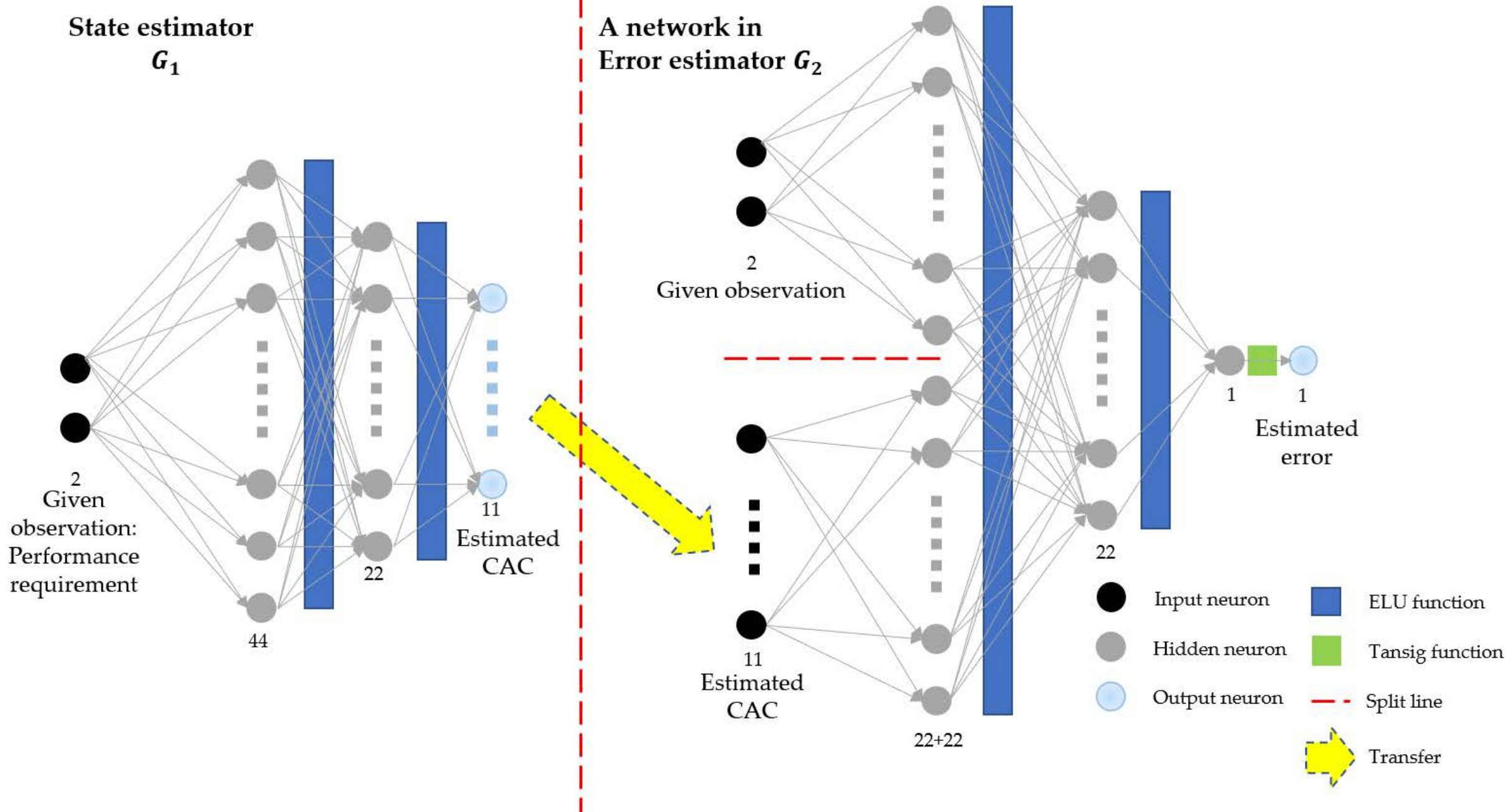}
\caption{ The architectures of networks for the application of gas turbine}
\label{fig20}
\end{figure}

We proposed a MLP architecture for $G_1$, which takes an expansion-convergence style, in order to first decompose the $y_{P,req}$ into more comprehensive features, and then transform them to eleven outputs, i.e., $\hat{x}_{CAC}$. This style is commonly seen in network design, such as the very classical LeNet ~\cite{lecunGradientbasedLearningApplied1998}, and more advanced, Fourier Neural Operator ~\cite{liFourierNeuralOperator2021}. Hereby, we expand the two input neurons to 44 neurons and gradually shrinks to eleven outputs with a layer of 22 neurons as the passage (see left part of Fig.~\ref{fig20}). The design of this MLP is simple, definitely, one can propose more complex architectures, or somehow optimize the redundant neurons/layers. Still, as what we talked before, we would like to emphasize the capability of the system of SVPEN, rather than that of a single network architecture.

As for the network in the error estimator $G_2$, it takes both given performance parameters, i.e., (1,1), and the estimated normalized eleven CAC parameters $\hat{x}_{CAC,norm}$ as inputs, and to mimic the error. This architecture is shown on the right part of Fig.~\ref{fig20}. Similar to the MAS application, two branches were designed to transform both types of inputs, which is realized by a dense layer with 22 neurons, then the encoded inputs were concatenated, and transformed to a scalar, i.e., the estimated error, through a hidden layer with 22 neurons.

\textbf{Physical forward model.} An open-source forward gas turbine engine model, Huracan \url{https://github.com/alopezrivera/huracan}, was used as the forward physical model. By using the parameters listed in  Table~\ref{Table2}, the thrust and TSFC were calculated as 126.83 KN and 11.22 g/(kN.s), respectively. These values include a relative error of 5\% in regards to the performance expected, so the modeling fidelity of Huracan is admissible.

\textbf{Error calculation component.} The purpose of this application study is through modifying the error calculation component to reflect different design requirements and thus demonstrating the reconfigurability of SVPEN. As what we summarized in Section ~\ref{3.1.2}. The physical evaluation module is not only the judge but also the guider.Thereby, given the rank-deficiency and ill-posedness of the problem at hand, we can utilize the second role of physical evaluation module by embedding design constraints into the error calculation component to further guide the optimization direction of the CAC parameters. The specific error calculation component design is discussed detailed in following subsections.

\subsubsection{SVPEN performance}\label{3.2.2}
In what follows, we present five studies to show the reconfigurability of SVPEN, all of which are directed to the optimization direction of SVPEN by regulating the error calculation method. They are:
\begin{itemize}
    \item (i) the cases of benchmark. In this case, we used the regularizations of technical solution costs and technology mismatch to direct the optimization direction
 \item (ii) the case of relaxing component parameters. Based on the case of benchmark, we almost eliminated the technical solution costs caused by using advanced component parameters.
 \item (iii) the case of relaxing technology matching. Based on the case of relaxing component parameters, we allowed certain technology mismatch between CAC parameters.
 \item (iv) the case of 20\% TSFC reduction. In this case, we further proposed a Turbofan design with 20\% less fuel consumption than the original CFM56-7B.
 \item (v) the case of unlocking CAC potential. In this case, we removed all regularizations of technical solution costs and technology mismatches, to assess the extreme performance can be offered by configured feasible domain of CAC. This case, in particular, demonstrates the expanded optimization mode of SVPEN, which was discussed in section ~\ref{2.3}. 
\end{itemize}

\textbf{(i) Benchmark:} Following the previous discussion on the error calculation component, we first considered the means to calculate the discrepancy between given and estimated observations, i.e., $F$ and $TSFC$. According to the physical meanings of $F$ and $TSFC$, it is inacceptable that the engine designed provides a smaller $F$ but higher $TSFC$ than the target values. Therefore, we defined acceptable normalized FN is 100\%-105\%, and acceptable normalized TSFC is 95\%-100\%, and the error $e_y$ can be defined as elaborated in Eqs.~\ref{eq35}-~\ref{eq37}:
\begin{equation}
\label{eq35}
    e_{F,1}=Max(1-\frac{\hat{F}}{F_{req}},0)+Max(\frac{\hat{F}}{F_{req}}-1.05,0)
\end{equation}
\begin{equation}
\label{eq36}
    e_{TSFC,1}=Max(0.95-\frac{\hat{TSFC}}{TSFC_{req}},0)+Max(\frac{\hat{TSFC}}{TSFC_{req}}-1.0,0)
\end{equation}
\begin{equation}
\label{eq37}
    e_{y,1}=\frac{e_{F,1}+e_{TSFC,1}}{2}
\end{equation}
Where, $e_{F,1}, e_{TSFC,1},e_{y,1}$,are respectively the discrepancy on $F$, the discrepancy on $TSFC$, and the average of these two terms. The subscript $1$ is used to demonstrate this is the first design of the formula, which is used to distiguish the current version from the modified version of these error terms in later cases. As elaborated in Eqs.~\ref{eq35}~\ref{eq36}, the 1 and 1.05 in Eq.~\ref{eq35} represents the 100\% and 105\% of normalized $F$; the 0.95 and 1 in Eq.~\ref{eq36} represent the 95\% and 100\% of normalized $TSFC$. Therefore, the $e_{y,1}$ defined leaves 5\% one-side tolerance margin for both $F$ and $TSFC$, the estimated $F$ and $TSFC$ dropping into this margin would not lead to extra error.

Similar to the MAS case, the feasible domain of the CAC parameters was utilized. The use of the feasible domain brought two-fold benefits.(1) It reduced the number of unnecessary attempts (2) It alleviated the problem of physically meaningless design, e.g., where the design parameters cause the pressure at the nozzle inlet to be smaller than that of the environment. As physically meaningless design may crash the physical forward model, we did not embed feasible domain into a regularization, instead, we compulsorily set the CAC parameters to be boundary values once they exceed the boundaries of feasible domain. The feasible domain of the CAC parameters was estimated according to ~\cite{kumarInnovativeApproachesReducing2012,kurzkePropulsionPowerExploration2018}, and demonstrated in  Table~\ref{Table3}.

\begin{table}
\centering
\caption{ Pre-defined feasible domain of CAC parameters}
\label{Table3}
\begin{tabular}{cccccccccccc} 
\hline
\textbf{Boundary} & \textbf{$BPR$} & \textbf{$\pi_{fan}$} & \textbf{$\pi_{LC}$} & \textbf{$\pi_{HC}$} & \textbf{$T_{max}$} & \textbf{$\eta_{fan}$} & \textbf{$\eta_{HC}$} & \textbf{$\eta_{LC}$} & \textbf{$\eta_{B}$} & \textbf{$\eta_{HT}$} & \textbf{$\eta_{LT}$}  \\ 
\hline
Lower             & 5          & 1.3        & 1.2        & 8          & 1300 K     & 0.85       & 0.82       & 0.84       & 0.95       & 0.86        & 0.87         \\
Upper             & 6          & 2.5        & 2          & 15         & 1800 K     & 0.95       & 0.92       & 0.94       & 0.995      & 0.96        & 0.97         \\ 
\hline
                  &            &            &            &            &            &            &            &            &            &             &              \\
                  &            &            &            &            &            &            &            &            &            &             &              \\
                  &            &            &            &            &            &            &            &            &            &             &              \\
                  &            &            & 1          &            &            &            &            &            &            &             &             
\end{tabular}
\end{table}

In addition, some pragmatic regularizations were utilized in order ot embody design preferences and alleviate the occurrence of ill-posed solutions. The regularizations utilized herein include technical solution costs and technology mismatch issues. The incorporation of technical solution costs is rather intuitive. As we know, more advanced technologies do not only bring better performance but also lead to higher costs. Therefore, from the perspective of design product, a common requirement/objective is to deliver the required functionality with minimal expenses. In the case at hand, the technical advance is embodied by higher CAC parameters, for example, 1300 K for $T_{max}$ is ordinary, but 1800 K requires much more sophisticated cooling design ~\cite{bunkerGasTurbineCooling2013} and the utilization of expensive ceramic thermal barrier coating ~\cite{padtureEnvironmentalDegradationHightemperature2019}. Therefore, the more advanced the CAC parameters, the more expensive is the associated solution. To appeal to this consideration, a regularization error term related to technical costs $e_{reg,TC}$ is defined as shown in Eq.~\ref{eq38}, which in fact the average of all CAC parameters, where $m=11$ herein. This regularization term is based on the arbitrary hypothesis that every CAC parameter has the same importance, which translates to that the more advanced CAC parameters will lead to higher error. Another term related to technology matching is also intuitive, i.e., the components should have similar technological levels. For instance, it is difficult to consider a CPU, which incorporates an advanced processor interfacing with I/O devices through analog components. The error regularization term related to technology matching is chosen as the standard derivation of the normalized CAC parameters, as shown in Eq.~\ref{eq39}. Consequently, the whole error can be defined as shown in Eq.~\ref{eq40}.
\begin{equation}
\label{eq38}
    e_{reg,TC,1}=\frac{\sum^{m}_{i=1}\hat{x}_{CAC,norm,i}}{m}
\end{equation}
\begin{equation}
\label{eq39}
    e_{reg,TM,1}=\frac{\|\hat{x}_{CAC,norm}-e_{reg,TC,1}\|_2}{m}
\end{equation}
\begin{equation}
\label{eq40}
    e_1=0.8e_{y,1}+0.1e_{reg,TC,1}+0.1 e_{reg,TM,1}
\end{equation}

The weights of the associated error terms can be considered as hyperparameters, which depend both on the importance and the magnitude of each term. The choice of these parameters can be further investigated using the paradigm of multi-task learning ~\cite{gongComparisonLossWeighting2019}.

\begin{figure}[hbt!]
\centering
\includegraphics[width=.6\textwidth]{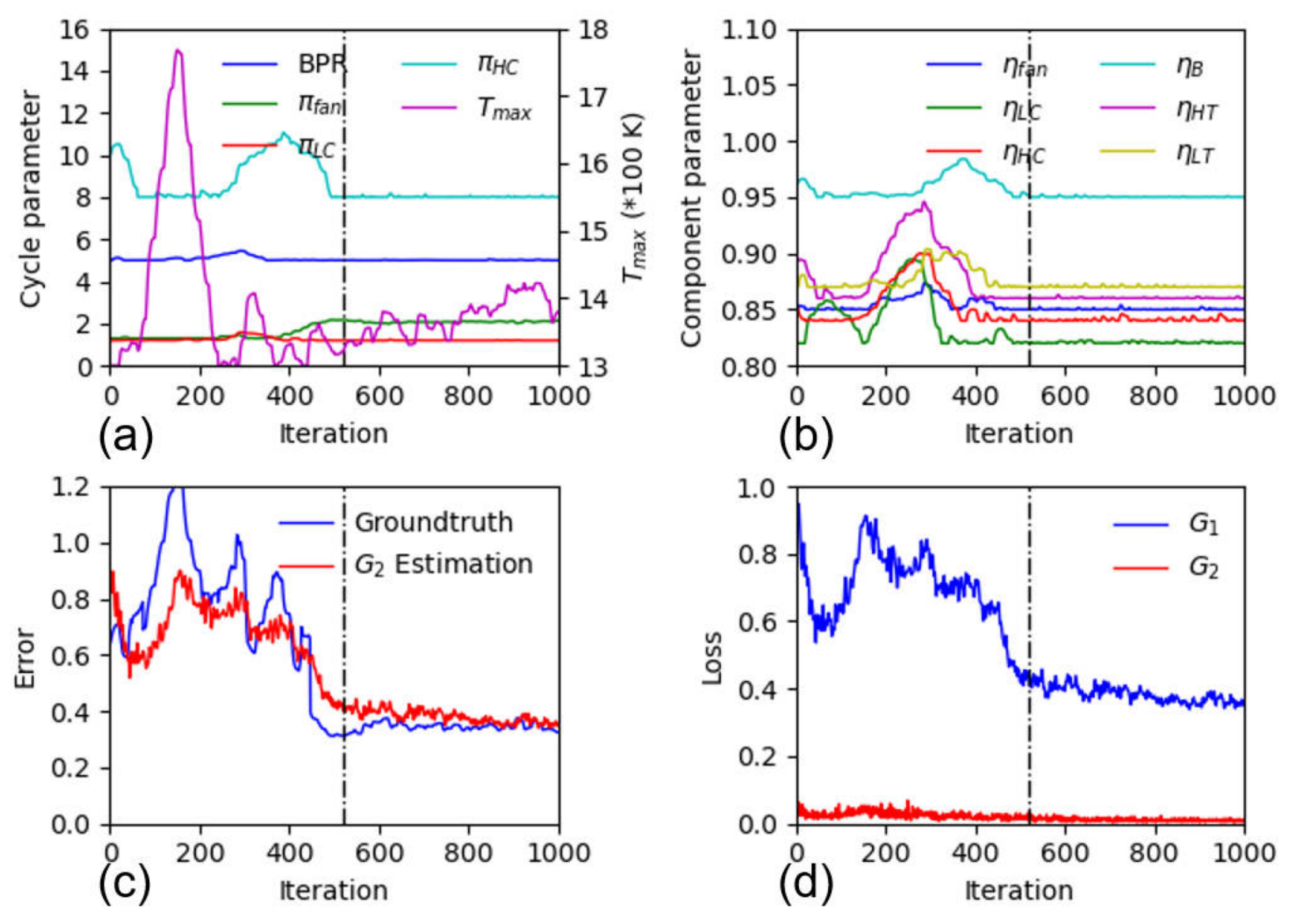}
\caption{ Details of the optimization process for the benchmark case. (a) estimated values of cycle parameters among the iterations, (b) estimated values of component parameters among the iterations, (c) error changes among the iterations, (d) Loss changes among the iterations.}
\label{fig21}
\end{figure}

As previously mentioned, only the optimization mode of SVPEN is available for this study. In the absence of ground truth of state to be used as a reference to set the error threshold empirically, SVPEN was ran for 1000 iterations instead, and the optimal CAC parameters was the ones corresponding to the lowest error among the iterations. The optimization process of the benchmark case is illustrated in Fig.~\ref{fig21}. As can be seen, only $T_{max}$ changes violently during the iteration, whereas the remaining CAC parameters almost keep constant in the last 500 iterations, the $BPR$ and $\pi_{LC}$ are even almost unchanged during the whole process. As for the update of both estimators, we can see that $G_2$ gradually learns the way to estimate the ground truth of error (Fig. ~\ref{fig21} (c)): it captures the trend and even several sudden peaks of the error. The reducing trend of estimated error also tells that $G_1$ works in a right way.

The lowest error is reached after 521 iterations, as pointed by the vertical black dash line in each subplot. The estimated CAC parameters corresponding to this error value are listed in the first column of  Table~\ref{Table4}: Apart from $\pi_{fan}$, all other CAC parameters are almost locked at the predefined lower boundary, especially, the component parameters. This state estimation is reasonable since the whole error imposes SVPEN to reduce both the magnitude and variance of the CAC parameters. In turn, this group of estimated CAC parameters also provides valuable physical insight, i.e., the performance is more sensitive to the cycle parameters than the component parameters, which is pretty obvious when looking at the change of $T_{max}$ and final estimation of $\pi_{fan}$. This is intuitive to understand, since the fuel, the source of $TSFC$, directly controls $T_{max}$, thus active fluctuation of $T_{max}$ is observed. Besides, Any increase of $\pi_{fan}$ will boost the momentum of all air sucked by the engine. This physical insight explains why researchers chose to fix component parameters or link them to cycle parameters ~\cite{liewParametricCycleAnalysis2005,mattinglyAircraftEngineDesign2018}, so as to simplify the complexities of cycle analysis.

\begin{table}
\centering
\caption{ Engine CAC parameters estimation}
\label{Table4}
\begin{tabular}{ccccccc} 
\hline
\multicolumn{2}{c}{\textbf{parameter}} & \textbf{Benchmark} & \begin{tabular}[c]{@{}c@{}}\textbf{Relaxing }\\\textbf{component }\\\textbf{parameters}\end{tabular} & \begin{tabular}[c]{@{}c@{}}\textbf{Relaxing }\\\textbf{technology}\\\textbf{~matching}\end{tabular} & \begin{tabular}[c]{@{}c@{}}\textbf{20\% }\\\textbf{TSFC }\\\textbf{reduction}\end{tabular} & \begin{tabular}[c]{@{}c@{}}\textbf{Unlocking}\\\textbf{~CAC}\\\textbf{~potential}\end{tabular}  \\ 
\hline
Estimated   & $BPR$                        & 5.014              & 5                                                                                                    & 5.0114                                                                                              & 5.093                                                                                      & 5.972                                                                                           \\
State       & $\pi_{fan}$                       & 2.179              & 2.183                                                                                                & 1.964                                                                                               & 1.965                                                                                      & 2.5                                                                                             \\
            & $\pi_{LC}$                        & 1.2                & 1.2                                                                                                  & 1.3573                                                                                              & 1.21                                                                                       & 2                                                                                               \\
            & $\pi_{HC}$                         & 8.001              & 8.019                                                                                                & 8                                                                                                   & 8.294                                                                                      & 15                                                                                              \\
            & $T_{max} (K)$                         & 1325               & 1315.6                                                                                               & 1618.7                                                                                              & 1300                                                                                       & 1301.66                                                                                         \\
            & $\eta_{fan}$                       & 0.85               & 0.851                                                                                                & 0.853                                                                                               & 0.95                                                                                       & 0.949                                                                                           \\
            & $\eta_{LC}$                       & 0.82               & 0.8259                                                                                               & 0.8839                                                                                              & 0.8212                                                                                     & 0.92                                                                                            \\
            & $\eta_{HC}$                         & 0.84               & 0.8425                                                                                               & 0.8415                                                                                              & 0.8809                                                                                     & 0.9398                                                                                          \\
            & $\eta_{B}$                        & 0.95               & 0.95                                                                                                 & 0.95                                                                                                & 0.989                                                                                      & 0.995                                                                                           \\
            & $\eta_{HT}$                        & 0.8606             & 0.8604                                                                                               & 0.9022                                                                                              & 0.9589                                                                                     & 0.9596                                                                                          \\
            & $\eta_{LT}$                        & 0.8704             & 0.8712                                                                                               & 0.8706                                                                                              & 0.9203                                                                                     & 0.9696                                                                                          \\
Estimated   & $F$ (kN)                       & 126.39             & 126.51                                                                                               & 130.023                                                                                             & 126.47                                                                                     & 180.27                                                                                          \\
observation & $TSFC$ (g/(kN.s))                       & 10.7               & 10.76                                                                                                & 11.17                                                                                               & 8.55                                                                                       & 8.82                                                                                            \\
\multicolumn{2}{c}{Optimal round}      & 521                & 504                                                                                                  & 516                                                                                                 & 984                                                                                        & 601                                                                                             \\
\hline
\end{tabular}
\end{table}

\textbf{(ii) Relaxing component parameters.} In order to encourage more active changes in component parameters, Eq.~\ref{eq38} was modified to Eq.~\ref{eq41}, and replaced $e_{reg,TC,1}$ by $e_{reg,TC,2}$ in whole error (Eq.~\ref{eq42}).
\begin{equation}
\label{eq41}
     e_{reg,TC,2}=\frac{0.99\sum_{i=1}^{6}\hat{x}_{C,norm,i}+0.01\sum_{j=1}^{5}\hat{x}_{\eta,norm,j}}{5+6}
\end{equation}
\begin{equation}
\label{eq42}
    e_2=0.8e_{y,1}+0.1e_{reg,TC,2}+0.1 e_{reg,TM,1}
\end{equation}

\begin{figure}[hbt!]
\centering
\includegraphics[width=.6\textwidth]{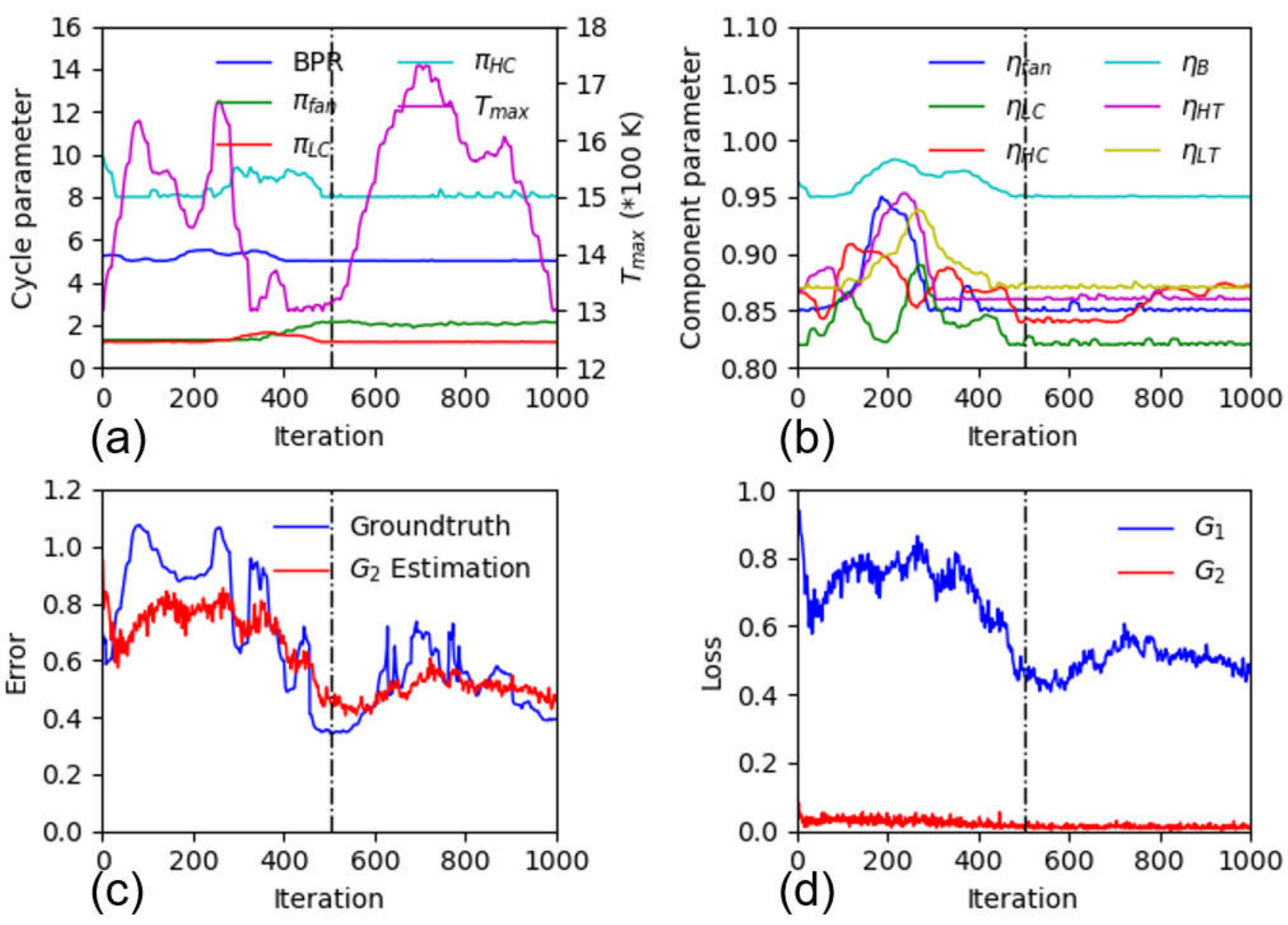}
\caption{ Details of the optimization process for the case of relaxing component parameters. (a) estimated values of cycle parameters among the iterations, (b) estimated values of component parameters among the iterations, (c) error changes among the iterations, (d) Loss changes among the iterations.}
\label{fig22}
\end{figure}

 As demonstrated by Eq.~\ref{eq41}, the weight of component parameters are neglectable comparing to that of cycle parameters, this means using advanced component parameters is “cheaper”, but using advanced cycle parameters is more “expensive”. However, this strategy does not work since the final estimated state has no significant difference to the former case (see  Table~\ref{Table4}), although there are more fluctuations in the iterations of the component parameters (as shown in Fig.~\ref{fig22}). The reason behind this is that the rigid technology matching regularization term (Eq.~\ref{eq39}) still bounds all CAC parameters.

\textbf{(iii) Relaxing technology matching.} In this investigation, we wish to consider the impact of the technology matching constraint. In this respect, it is unnecessary to strictly force all CAC parameters to be at the exact same technical level. Instead, we relax this regularization term as shown in Eq.~\ref{eq43}, and provide the corresponding total error in Eq.~\ref{eq44}.
\begin{equation}
\label{eq43}
    e_{reg,TM,2}=Max(e_{reg,TM,1}-0.4,0)
\end{equation}
\begin{equation}
\label{eq44}
    e_3=0.8e_{y,1}+0.1e_{reg,TC,2}+0.1 e_{reg,TM,2}
\end{equation}

\begin{figure}[hbt!]
\centering
\includegraphics[width=.6\textwidth]{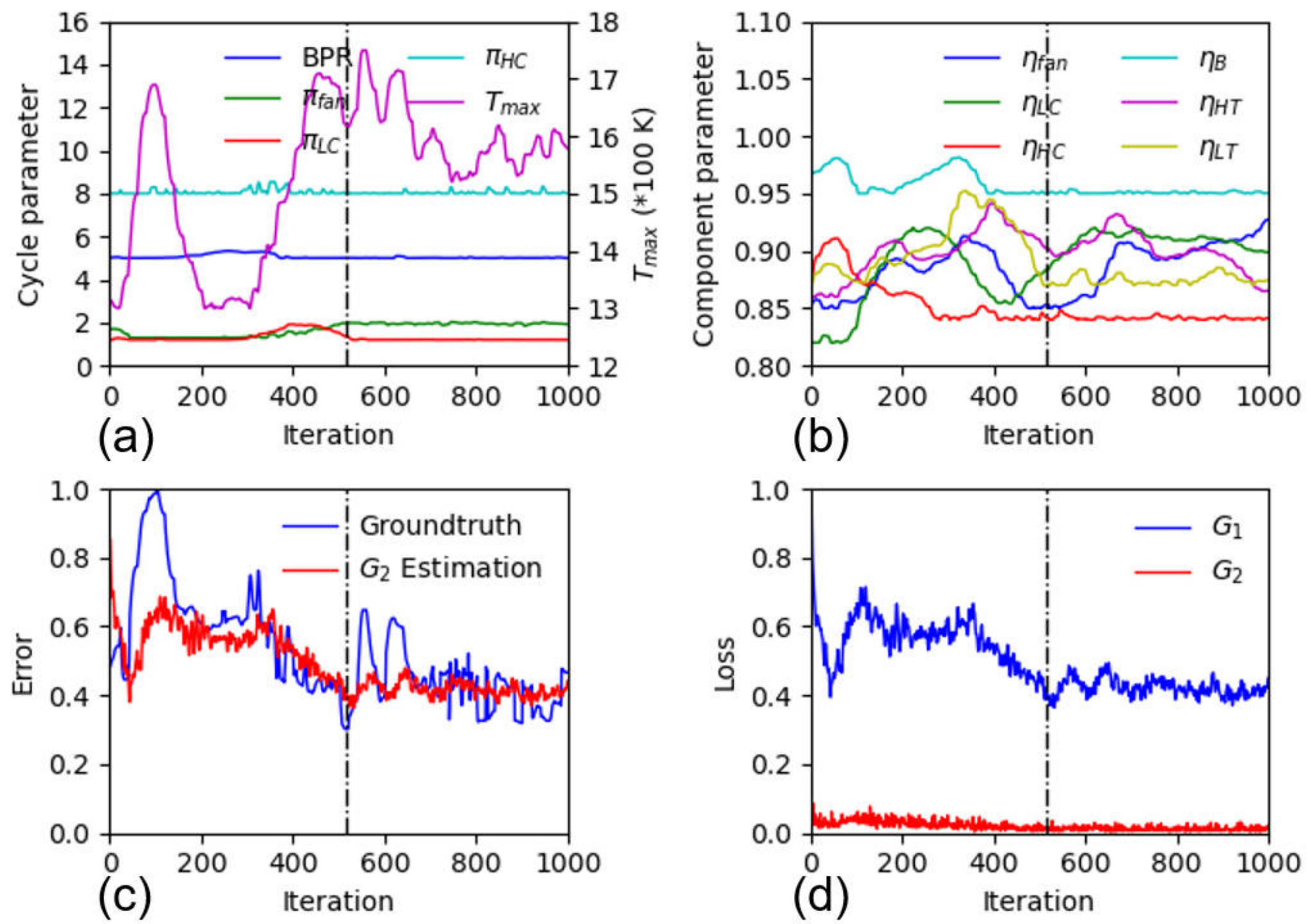}
\caption{ Details of the optimization process for the case of relaxing technological matching. ((a) estimated values of cycle parameters among the iterations, (b) estimated values of component parameters among the iterations, (c) error changes among the iterations, (d) Loss changes among the iterations.}
\label{fig23}
\end{figure}

As elabroated in Eq.~\ref{eq39}, the technology mismatch regularation only adds extra error when $e_{reg,TM,1}$ exceeds 0.4. As the definition of $e_{reg,TM,1}$ is standard error of normalized CAC parameters, 0.4 is a rather loose threshold. The reason of doing so is to highlight how error calculation significantly changes the direction of the optimization. Of course, one can choose looser or stricter threshold. With such an error function guiding the optimization, as shown in Fig.~\ref{fig23}, all component parameters are thoroughly activated. The final estimation shown in  Table~\ref{Table4} also tells that the values of $\eta_{LC}$ and $\eta_{HT}$ increase substantially compared to those in two previous cases, and the change in $T_{max}$ is significantly enhanced as well.

\textbf{(iv) 20\% TSFC reduction.} As CFM56 is an extremely mature product, following the trend of green aviation ~\cite{hughesAircraftEngineTechnology2011} , current engineers perhaps have more interest in providing a more economical engine at the same thrust level, e.g., an engine with 20\% less $TSFC$. To reflect such an design requirement, we redefined the error function as shown in Eqs.~\ref{eq45}- ~\ref{eq47}. The 0.75 and 0.8 in Eq.~\ref{eq45} respectively represent 75\% and 80\% of the $TSFC$ of CFM56, we still leave 5\% one-side margin as same as in Eq.(38), the $TSFC$ falls into this range will not be punished.
\begin{equation}
\label{eq45}
    e_{TSFC,2}=Max(0.75-\frac{\hat{TSFC}}{TSFC_{req}},0)+Max(\frac{\hat{TSFC}}{TSFC_{req}}-0.8,0)
\end{equation}
\begin{equation}
\label{eq46}
    e_{y,2}=\frac{e_{F,1}+e_{TSFC,2}}{2}
\end{equation}
\begin{equation}
\label{eq47}
    e_4=0.8e_{y,2}+0.1e_{reg,TC,2}+0.1 e_{reg,TM,2}
\end{equation}

\begin{figure}[hbt!]
\centering
\includegraphics[width=.6\textwidth]{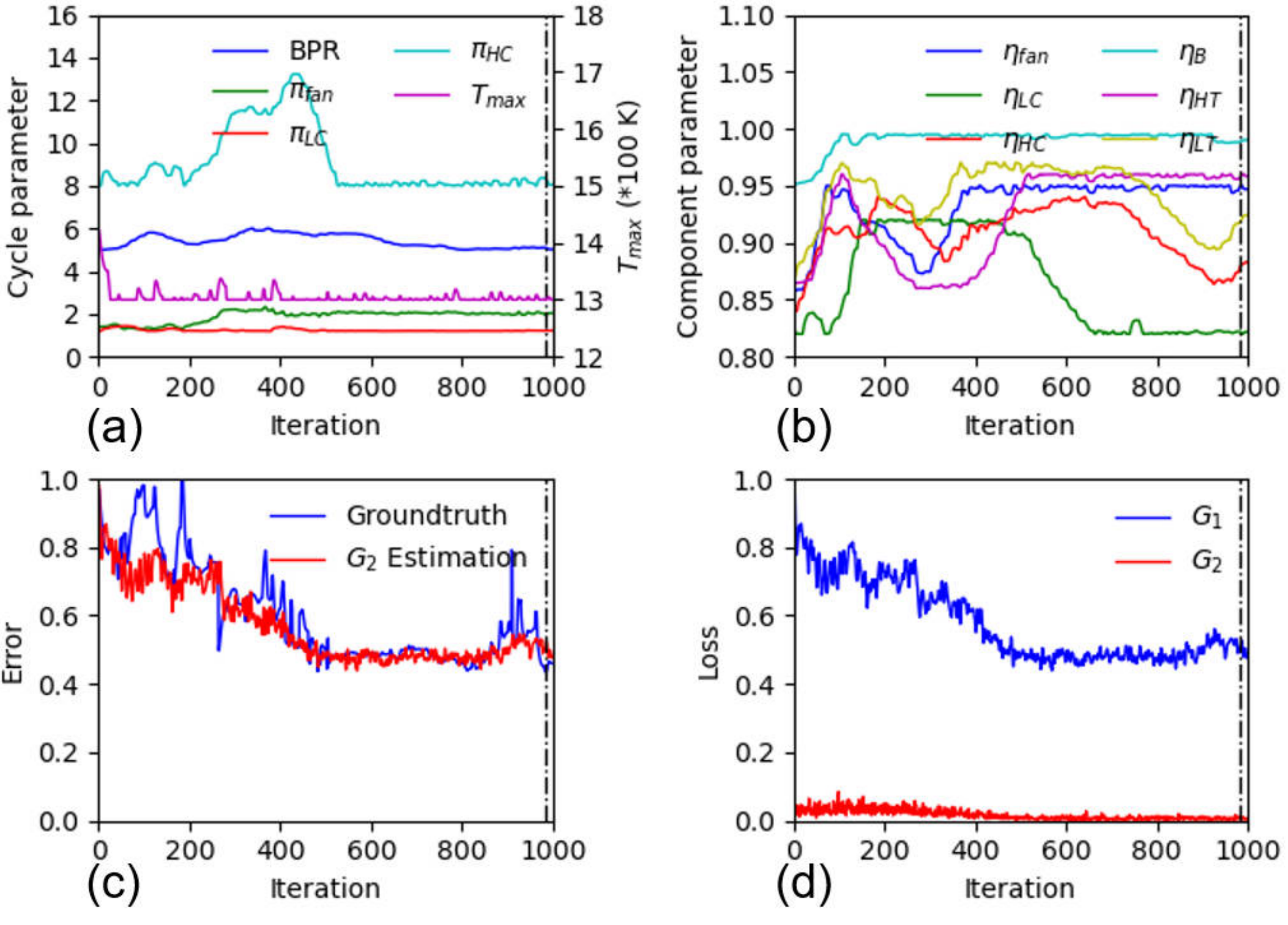}
\caption{ Details of the optimization process for the 20\% TSFC reduction case. (a) estimated values of cycle parameters among the iterations, (b) estimated values of component parameters among the iterations, (c) error changes among the iterations, (d) Loss changes among the iterations.}
\label{fig24}
\end{figure}

Fig. ~\ref{fig24} displays the optimziation process for this case. As can be seen, with such a stricter performance requirement, the design space is reduced, which is represented by that the large changes of the $T_{max}$ disappears, and instead, it is tightly bounded to the lower boundary of the feasible domain. This phenomenon is rationale as decreasing fuel consumption is the most straightforward way to decrease $TSFC$, the change of which is reflected on $T_{max}$. However, the decrease of fuel also causes $F$ loss, as one can imagine but unacceptable. Therefore, in order to maintain the $F$, one can also observed that most of component parameters climb to their upper boundaries along with the decrease of $T_{max}$. These improvements of component parameters are in fact a complement to reduction of fuel, so as to using the limited fuel with a higher efficiency. The optimal estimates of CAC parameters are met after 982 iterations, with the corresponding parameters summarized in  Table~\ref{Table4}.

\textbf{(v) Unlocking CAC potential.} The last investigation is to find the performance boundary that could be realized by the configured feasible domain of CAC parameters. To embody this idea, we removed the regularizations related to technology costs and mismatch, and modified the error function as elaborated by Eqs.~\ref{eq48}-~\ref{eq50}. The 1 in Eqs.~\ref{eq48}~\ref{eq49} are represent the self-normalized FN and TSFC of CFM56, so these two formulas conveys the information that the higher of $F$ and the lower of $TSFC$ with the reference of CFM56 is appreciated. Because the performance of CFM56 is just a reference, we changed $F_{req}$ and $TSFC_{req}$ used in former cases to $F_{ref}$ and $TSFC_{ref}$.
\begin{equation}
\label{eq48}
    e_{F,2}=exp(1-\frac{\hat{F}}{F_{ref}})
\end{equation}
\begin{equation}
\label{eq49}
    e_{TSFC,3}=exp(\frac{\hat{TSFC}}{TSFC_{ref}}-1)
\end{equation}
\begin{equation}
\label{eq50}
    e_{5}=\frac{e_{F,2}+E_{TSFC,3}}{2}
\end{equation}

It is notable that although we used (1,1) as the input to $G_1$, however, we do not know the exact values of observation, i.e., the exact value of $F_{req}$ and $TSFC_{req}$. Therefore, (1,1) here does not have any physical meaning, and can be replaced by any random vector. Till now, one can realize that this case has thorough difference from previous ones, as we only knows the property of wanted observation (Eqs.~\ref{eq48}~\ref{eq49}) but do not know the exact value of it, and thus, SVPEN works in the extended optimization mode (Fig~\ref{fig8}). 

\begin{figure}[hbt!]
\centering
\includegraphics[width=.6\textwidth]{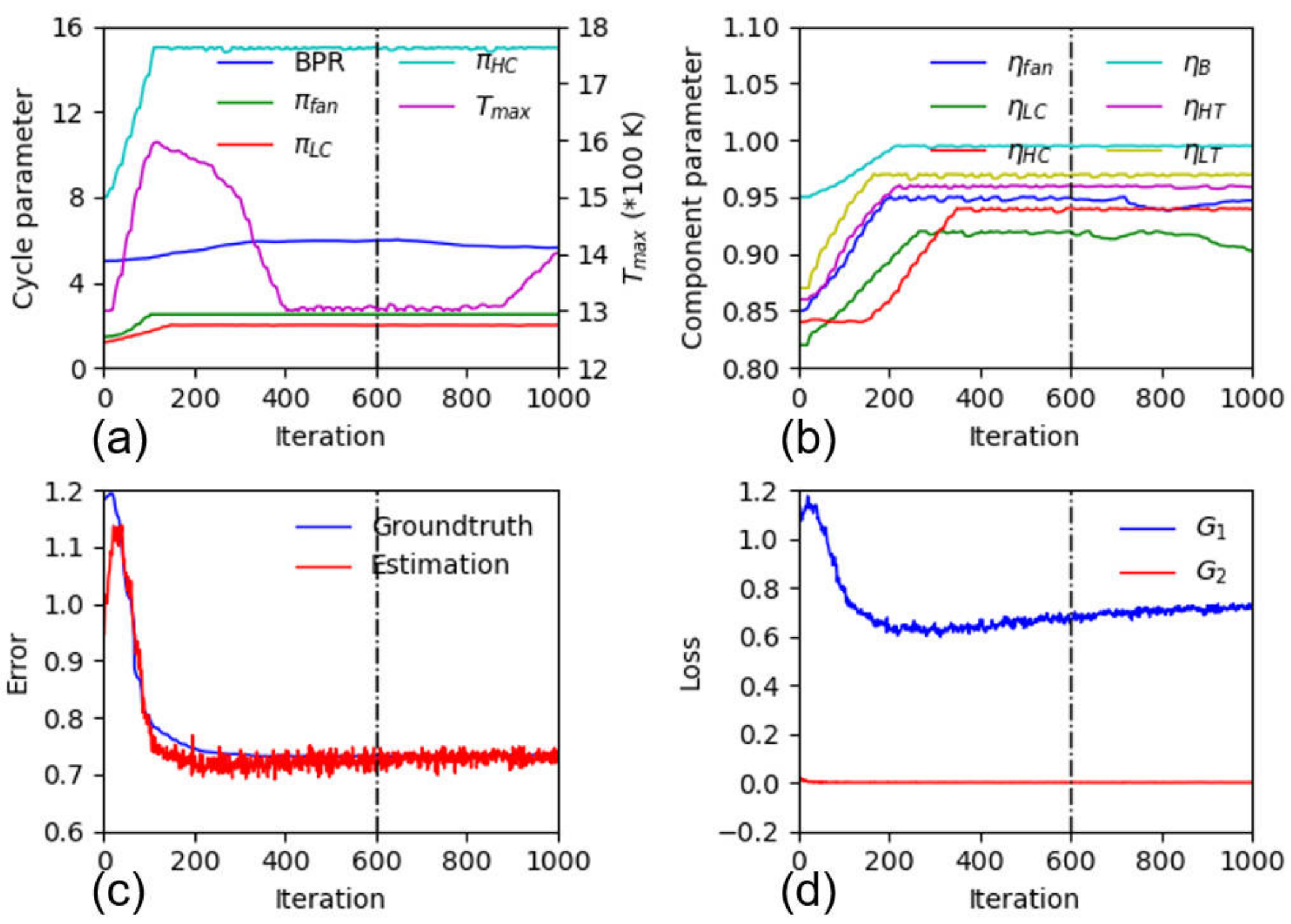}
\caption{ Detailed optimization process for the case of unlocking CAC potential. (a) estimated values of cycle parameters among the iterations, (b) estimated values of component parameters among the iterations, (c) error changes among the iterations, (d) Loss changes among the iterations.}
\label{fig25}
\end{figure}

Without any regularization terms, the power of this feasible domain is completely unlocked: all parameters except the maximum temperature are quickly lifting to the upper boundary, while the maximum temperature is diving to lower boundary ( Fig.~\ref{fig25}). Although, the minimal error is reached by 601 runs, in fact, this optimization process is almost converged after about 400 iterations. Both $G_1$ and $G_2$ learns very smoothly, since the estimated error is almost overlapped with the ground truth, and there is no fluctuation along with the endeavors of $G_1$ to decrease the error.

 Table~\ref{Table4} tells the same information that all parameters almost reach the upper boundaries except for the maximum temperature reaches the lower boundary. This result is apparently physically reasonable, because both higher pressure ratio and component parameters will lead to an higher cycle efficiency; the higher BPR will improve the propulsion efficiency; the lower maximum temperature, again, directly decrease the fuel consumption ~\cite{kurzkePropulsionPowerExploration2018}. The optimal $F$ and $TSFC$ are respectively 180.27 kN and 8.82 g/(kN.s), which is a significant improvement against the baseline performance: 126.83 kN and 11.22 g/(kN.s).

\section{Summary and conclusion}\label{4summary}
This work proposed Self-Validated Physics-Embedding Network (SVPEN), a general framework for inverse modelling. As the name suggests, this framework embeds a physical forward model to validate the quality of state estimation, and thus assure the acceptable estimation is physical reasonable. This crucial property cannot be assured by current pure data-driven supervised-learning-based models, since which is only "supervised" in training set, but works in an "unsupervised" way. 

The proposed framework has two work modes: inverse function mode and optimization mode. The inverse function provides the efficient solution through a neural network (state estimator) as what the conventional supervised learning. However, since the test observation cannot be always assured picked from the I.I.D of the trainig set of state estimator, the estimation of state cannot be always accurate,instead, may be very biased. Under this condition, the optimization mode of SVPEN offers the way that using another network, termed error estimator, to solve inverse problems via gradient-based optimization. The existence of optimization mode does not only provide the way of iterating biased solution, but also offers unprecedented reconfigurability of SVPEN, i.e., the components inside SVPEN, such as networks, physical models and error functions can be replaced at will, so that SVPEN can solve various inverse problems without collection of corresponding datasets and pretraining of state estimator. The key component, physical evaluation module, which wraps physical forward model and error calculation component, is not only the judge that assesses the quality of every estimated state, but also the guider that directs the optimization direction of optimization mode. 

Although more than ten case studies from two distinct application scenarios have demonstrated the reconfigurability and capability of SVPEN, we do understand the infinite and diverse inverse modelling problems would propose challenges to SVPEN, but we believe many of them can be solved by good formulation of the problem and utilizing the idea of SVPEN. What we want to emphasize is SVPEN is a general framework for inverse problems, so that it can help liberate researchers from sophisticated or even impossible physical inverse modelling. More importantly, SVPEN opens a window of exploiting the tremendous precious physical models, which have been developed and accumulated by uncounted intelligent researchers, thus, realizing a harmony cooperation between physical models and data-driven models.

\bibliographystyle{unsrt}  
\bibliography{reference}

\end{document}